\documentclass[twoside]{article}

\usepackage{graphicx, lipsum}
\usepackage{subfigure}
\usepackage{booktabs} % for professional tables
\usepackage{amsmath,amsfonts,bm}
\usepackage{xcolor}
\usepackage[normalem]{ulem}

\def\eqref#1{(\ref{#1})}
\def\1{\bm{1}}

\def\Dis{{\mathcal{D}is}}

\def\rvepsilon{{\mathbf{\epsilon}}}

\def\rvx{{\mathbf{x}}}
\def\rvy{{\mathbf{y}}}
\def\rvz{{\mathbf{z}}}

\def\vmu{{\bm{\mu}}}
\def\vtheta{{\bm{\theta}}}
\def\vvarphi{{\bm{\varphi}}}

\def\vvardelta{{\bm{\delta}}}
\def\va{{\bm{a}}}
\def\vb{{\bm{b}}}

\def\vs{{\bm{s}}}

\def\mA{{\bm{A}}}

\def\mI{{\bm{I}}}

\def\mM{{\bm{M}}}

\def\mU{{\bm{U}}}

\def\mLambda{{\bm{\Lambda}}}
\def\mSigma{{\bm{\Sigma}}}

\DeclareMathAlphabet{\mathsfit}{\encodingdefault}{\sfdefault}{m}{sl}
\SetMathAlphabet{\mathsfit}{bold}{\encodingdefault}{\sfdefault}{bx}{n}

\def\gD{{\mathcal{D}}}
\def\gE{{\mathcal{E}}}

\def\gI{{\mathcal{I}}}

\def\gN{{\mathcal{N}}}

\def\gQ{{\mathcal{Q}}}

\def\sR{{\mathbb{R}}}

\newcommand{\E}{\mathbb{E}}

\newcommand{\R}{\mathbb{R}}

\newcommand{\norm}[1]{\left\lVert#1\right\rVert}
\newcommand{\T}{\textup{T}}

\usepackage{xcolor}
\newtheorem{proposition}{Proposition}[section]
\usepackage{algorithm, algorithmic}
\usepackage{amsmath,amssymb}
\usepackage{xurl}

\usepackage{enumitem}
\usepackage[accepted]{aistats2023_cameraready}
\usepackage[round]{natbib}

\usepackage{xr}
\makeatletter
\newcommand{\addFileDependency}[1]{% argument=file name and extension
  \typeout{(#1)}
  \@addtofilelist{#1}
  \IfFileExists{#1}{}{\typeout{No file #1.}}
}
\makeatother

\newcommand{\myexternaldocument}[1]{%
    \externaldocument{#1}%
    \addFileDependency{#1.tex}%
    \addFileDependency{#1.aux}%
}

\myexternaldocument{supplement}

\AtEndDocument{\refstepcounter{equation}\label{finalequation}}
\AtEndDocument{\refstepcounter{figure}\label{finalfigure}}

\makeatletter
\newcommand{\printfnsymbol}[1]{%
  \textsuperscript{\@fnsymbol{#1}}%
}
\begin{document}



\twocolumn[

\aistatstitle{Robust Variational Autoencoding with Wasserstein Penalty for Novelty Detection}

\aistatsauthor{Chieh-Hsin Lai$^*$ \And Dongmian Zou$^*$ \And  Gilad Lerman }

\aistatsaddress{ School
of Mathematics, \\University of Minnesota \And  Division of Natural and Applied Sciences, \\Duke Kunshan University \And School
of Mathematics, \\University of Minnesota } 
% \thanks{$*$ indicates equal contribution.}
]

\begin{abstract}
We propose a new method for novelty detection that can tolerate high corruption of the training points, whereas previous works assumed either no or very low corruption. Our method trains a robust variational autoencoder (VAE), which aims to generate a model for the uncorrupted training points. To gain robustness to high corruption, we incorporate the following four changes to the common VAE: 1.~Extracting crucial features of the latent code by a carefully designed dimension reduction component for distributions; 2.~Modeling the latent distribution as a mixture of Gaussian low-rank inliers and full-rank outliers, where the testing only uses the inlier model; 3.~Applying the Wasserstein-1 metric for regularization, instead of the Kullback-Leibler (KL) divergence; and 4.~Using a 
robust error for reconstruction. We establish both robustness to outliers and suitability to low-rank modeling of the Wasserstein metric as opposed to the KL divergence. We illustrate state-of-the-art results on standard benchmarks. 
\end{abstract}

\section{INTRODUCTION}
\label{sec:intro}
Machine learning solutions often assume that training datasets are flawless and can serve as ground truth. However, this assumption usually does not hold in practice. Indeed, most datasets, even commonly used ones such as CIFAR-10 or ImageNet, suffer from corruption and mislabeling \citep{northcutt2021pervasive}. While in many applications the percentage of mislabels may be sufficiently small, there are important scenarios where this is not the case. One such scenario appears when studying problems with no earlier experience and expertise. For instance, in the beginning of the COVID-19 pandemic it was hard to diagnose COVID-19 patients and distinguish them from other patients with pneumonia \citep{chowdhury2020can, xiao2020false}. 
Another scenario occurs when it is very hard to make precise measurements, for example, when working with the highly corrupted images in cryogenic electron microscopy (cryo-EM) \citep{miolane2020estimation, huang2015robust}.

One problem, where it is crucial to carefully address mislabeled training data points, is novelty detection. It asks to  
detect testing data points that deviate from the underlying structure of a given training dataset \citep{chandola2009anomaly, pimentel2014review, chalapathy2019deep, perera2021one}. 
% Ideally, it requires learning the underlying distribution of the training data, where sometimes it is sufficient to learn a significant feature, geometric structure or another property of the training data. One can then apply the learned distribution (or property) 
% to detect deviating points in the test data. 
Novelty detection is equivalent to the well-known one-class classification problem \citep{moya1996network}. 
This problem asks to identify members of a class in a test dataset, and consequently distinguish them from ``novel'' data points, given training points from this class. The points of the main class are commonly referred to as inliers and the novel ones as outliers. Novelty detection is also commonly referred to as semi-supervised anomaly detection. In this terminology, the notion of being ``semi-supervised'' is different from usual, and means that a training set is provided for the inliers only. 
On the other hand,  the supervised case has labeled training data for both the inliers and outliers, and the unsupervised case has no training and is also known as  ``outlier detection''.

Traditional one-class classification methods often assume that the training set is purely sampled from a single class or has few outliers and perform poorly when there is a nontrivial portion of outliers. In this paper, we study a robust version of novelty detection that allows a nontrivial fraction of corrupted samples, namely outliers, within the training set.
We solve this problem by using a special variational autoencoder (VAE) \citep{kingma2014auto}. Our VAE is able to model the underlying distribution of the uncorrupted data, despite nontrivial corruption.  
We refer to it as ``Mixture Autoencoding with Wasserstein penalty'', or ``MAW''. 
% In order to clarify it, we first review previous works and then explain our contributions in view of these works. 

\subsection{Previous Work}\label{sec:related_work}

Solutions to novelty detection either estimate the density of the inlier distribution \citep{bengio2005non, ilonen2006gaussian} or determine a geometric property of the inliers, such as their boundary set \citep{breunig2000lof, scholkopf2000support, xiao2016robust, wang2020robust, jiang2019fast}.
When the inlier distribution is nicely approximated by a low-dimensional linear subspace,  \citet{shyu2003novel}
propose to distinguish between inliers and outliers via Principal Component Analysis (PCA). 
In order to consider more general cases of nonlinear low-dimensional structures, one may use autoencoders (or restricted Boltzmann machines), which nonlinearly generalize PCA~\citep[Ch.~2]{goodfellow2016deep} and whose reconstruction error naturally provides a score for membership in the inlier class. Instances of this strategy with various architectures include \citep{zhai2016deep, zong2018deep, sabokrou2018adversarially, perera2019ocgan, pidhorskyi2018generative}. 
In all of these works, but \citep{zong2018deep}, the training set is assumed to solely represent the inlier class. 
If there are also outliers (with a simple shape) among the inliers (with a complex shape), encoding the inlier distribution becomes difficult. Nevertheless, some previous works already explored the possibility of a corrupted training set \citep{xiao2016robust, wang2020robust, zong2018deep}. In particular, \cite{xiao2016robust, zong2018deep} test artificial instances with at most $5\%$ corruption of the training set and \cite{wang2020robust}
consider ratios of $10\%$, but with very small numbers of training points. In this work we consider corruption ratios up to $50\%$, with a method that tries to estimate the distribution of the training set, and not just a geometric property. 

% One may model the low-dimensional structure by several modes \citep{zong2018deep, fan2020correlation} in order to better capture inhomogeneous characteristics of the inliers; such modeling also appears in other machine learning tasks \citep{zhang2020one, lu2020semantic}. Nevertheless, the bi-modal latent model in this work is designed to distinguish between the inliers and outliers. We believe that looking for additional modes of inliers may rather complicate our model and may distract it from detecting outliers.

VAEs \citep{kingma2014auto} have been commonly used for generating distributions with reconstruction scores and are thus natural for novelty detection without corruption. 
% They determine the latent code of an autoencoder via variational inference \citep{jordan1999introduction, blei2017variational}. Alternatively, they can be viewed as autoencoders for distributions that penalize the Kullback-Leibler (KL) divergence of the latent distribution from the prior distribution.
The first VAE-based method for novelty detection was suggested by \citet{an2015variational}. It was recently extended by \citet{daniel2020deep} who modified the training objective. 
A variety of VAE models were also proposed for special anomaly detection problems, which are different from novelty detection~\citep{xu2018unsupervised,zhang2019velc,pol2019anomaly}. Current VAE-based methods for novelty detection do not perform well when the training data is corrupted. Indeed, the learned distribution of any such method also represents  corruption, that is, the outlier component. To the best of our knowledge, no effective solutions were proposed for collapsing the outlier mode so that the trained VAE would only represent the inlier distribution.

A variant of VAE is the adversarial autoencoder (AAE) of \citep{makhzani2015adversarial}.
The penalty term of AAE takes the form of a generative adversarial network (GAN) \citep{goodfellow2016deep}, where the AAE's encoder serves as the GAN's generator. We can thus view it as a hybrid GAN-VAE model. 
Another such model is the Wasserstein autoencoder (WAE) \citep{tolstikhin2018wasserstein}, which generalizes AAE by allowing a general objective function. Our proposed model is also a hybrid GAN-VAE. Other 
hybrid VAE-GAN models include \citep{mescheder2017adversarial, xian2019f, ye2021infovaegan}. The GAN of \citep{mescheder2017adversarial} is used for both the samples and the latent code, 
the GAN of \citep{xian2019f, ye2021infovaegan} is used only for the samples, whereas the GAN of our work and \citep{makhzani2015adversarial, tolstikhin2018wasserstein} is used only for the latent code. We demonstrate the resulting robustness to outliers due to our particular use of a GAN.

There are two relevant lines of work on robustness to outliers in linear modeling that can be used in nonlinear settings via autoencoders or VAEs. 
Robust PCA aims to deal with sparse elementwise corruption of a data matrix  \citep{candes2011robust,Torre:03,wright2009robust,vaswani2018static}. Robust subspace recovery (RSR) aims to address general corruption of selected data points and thus better fits the framework of outliers 
\citep{watson2001some, Torre:03, ding2006r, zhang2009median, mccoy2011two, xu2012robust, lp_recovery_part1_11, zhang2014novel, lerman2015robust, lerman2017fast, maunu2019well, lerman2018overview, maunu2019robust}.
Autoencoders that use robust PCA for anomaly detection tasks were proposed in \citep{chalapathy2017robust, zhou2017anomaly}. 
It is shown in \citep{dai2018connections} that a VAE can be interpreted as a nonlinear robust PCA problem. Nevertheless, explicit regularization is often required to improve robustness to sparse corruption in VAEs~\citep{akrami2019robust,eduardo2020robust}.
An RSR layer was successfully applied to outlier detection in \citep{lai2020robust}. One can also apply this work to novelty detection. 

We remark that the setting of our work is different from that of out-of-distribution (OOD) detection and open-set recognition. Indeed, in these recent settings the inliers are from multiple classes that need to be identified. On the other hand, this work does not ask to classify the inliers.

\subsection{This Work}\label{subsec:contribution}

We propose a robust novelty detection procedure, MAW, that aims to model the distribution of the training data in the presence of a nontrivial fraction of outliers. We highlight its following four features:
\begin{enumerate}%[leftmargin=12pt, parsep=0pt]
\item
MAW models the latent distribution by a Gaussian mixture of low-rank inliers and full-rank outliers, and applies the inlier distribution for testing. Previous applications of mixture models for novelty detection were designed for multiple modes of inliers and used more complicated tools such as additional network construction \citep{zong2018deep} or clustering \citep{aytekin2018clustering,lee2018simple}.
\item MAW applies a novel dimension reduction component, which extracts lower-dimensional features of the latent distribution. The reduced dimension allows using full covariances; whereas previous VAE-based methods for novelty detection used diagonal covariances in their models \citep{an2015variational, daniel2020deep}. 
%The new component is inspired by the RSR layer in \citep{lai2020robust}; however, they are essentially different since the RSR layer only applies to data points and not to distributions.
\item
MAW uses the Wasserstein-1 ($W_1$) metric for the latent code penalty. We prove that the Wasserstein metric gives rise to outlier-robust estimation and is suitable to the low-rank modeling of inliers by MAW. We also show that these properties do not hold for the commonly-used KL divergence. To the best of our knowledge, this is the first theoretical analysis that clarifies the advantage of the Wasserstein distance over the KL divergence in a VAE in terms of robustness to outliers and low-rank inlier modeling. 
%We remark that the use of $W_1$ in WAE is different from that of MAW. Indeed, in WAE $W_1$ measures the distance between the data distribution and the generated distribution and it does not appear in the latent code.
%Our use of $W_1$ can be viewed as a variant of AAE, which replaces GAN with Wasserstein GAN (WGAN)~\citep{arjovsky2017wasserstein}, and thus replaces the minimization of the KL divergence by that of the $W_1$ distance.
\item
MAW achieves state-of-the-art results on popular anomaly detection datasets.
\end{enumerate}

Additional two features are as follows. First, for reconstruction, MAW replaces the common least squares formulation with a least absolute deviations formulation. This can be justified by the use of a robust estimator~\citep{lopuhaa1991} with a heavier-tail likelihood.
Second, MAW is attractive for practitioners. It is simple to implement in any standard deep learning library, and is easily adaptable to other choices of network architecture, energy functions and similarity scores.

\section{DESCRIPTION OF MAW}
\label{sec:method}

We motivate and overview the underlying model and assumptions of MAW in \S\ref{subsec:motivation_overview}. We describe the implementation details of its components in \S\ref{subsec:details}
and sketch the algorithm procedures in  
the supplementary materials. 
Fig.~\ref{fig:arch} illustrates the general idea of MAW and can assist in reading this section.

\begin{figure*}[!th]
  \centering
  \includegraphics[width=0.9\textwidth]{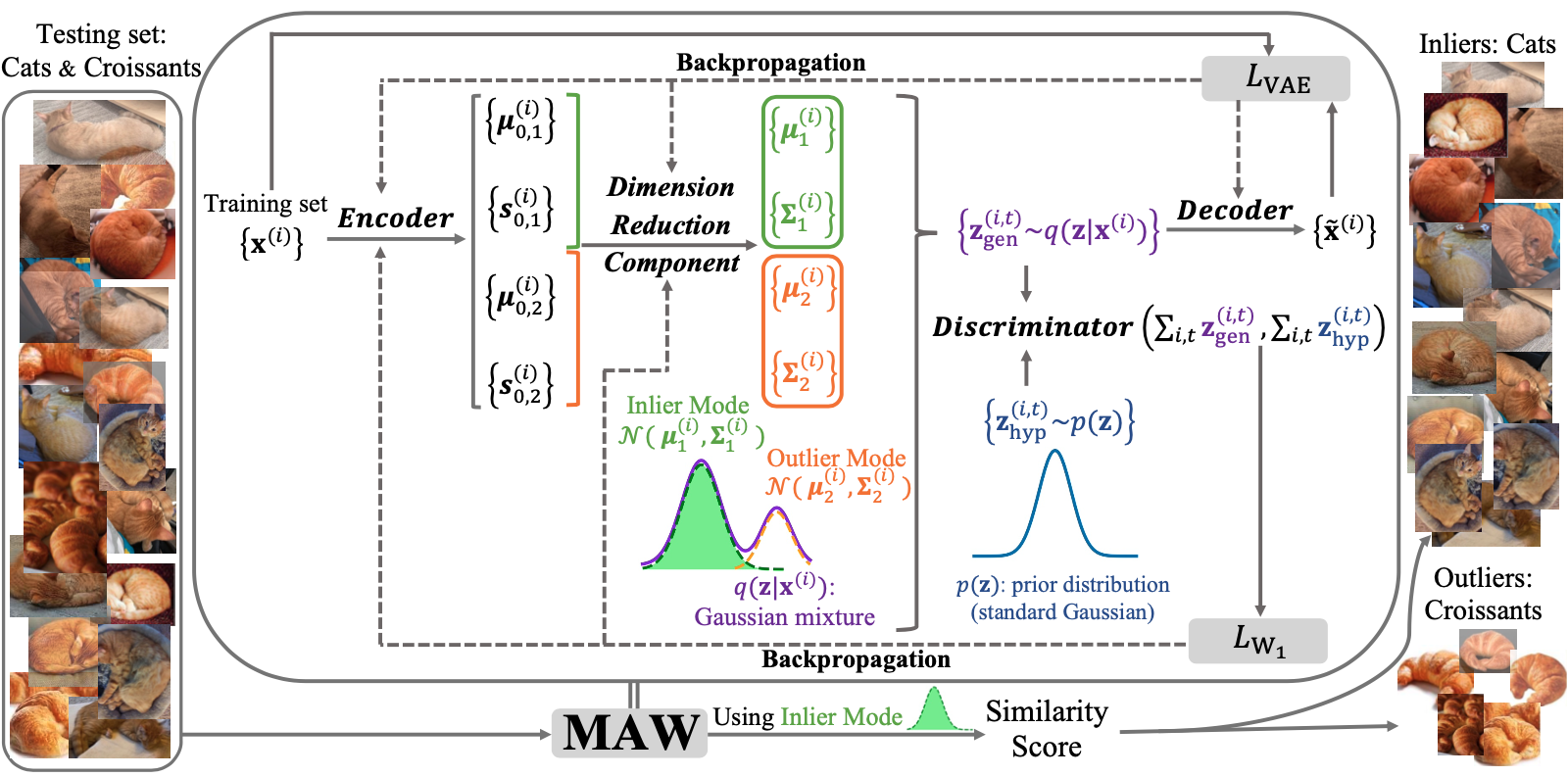}
  \begin{center}
  \caption{Demonstration of the architecture of MAW for novelty detection. \label{fig:arch}}       
  \end{center}
  \vskip -0.2in
\end{figure*}

\subsection{The Model and Assumptions of MAW}\label{subsec:motivation_overview}

MAW aims to robustly estimate a mixture inlier-outlier distribution
for the training data and then use its inlier component to detect
outliers in the testing data. For this purpose, it designs a novel variational autoencoder with an underlying mixture model and a robust loss function in the latent space. We find the variational framework natural for novelty detection. Indeed, it learns a distribution that describes the inlier training examples and generalizes to the inlier test data. Moreover, the variational formulation allows a direct modeling of a Gaussian mixture model (GMM) in the latent space, unlike a standard autoencoder.

Let $\rvx$ be a random variable in $\R^D$ with an unknown training data distribution, which contains both inlier and outlier modes. We assume $L$ training points in $\R^D$, $\{\rvx^{(i)}\}_{i=1}^L$ sampled from this distribution. 
We assume a latent random variable $\rvz$ of low and even dimension $2 \leq d \leq D$ (our default choice is $d=2$), and a standardized Gaussian prior, $p(\rvz)$, so that $\rvz \sim \gN(\bm{0}, \mI_{d \times d})$. In the remaining text, we shall denote it as 
\begin{equation}
\nonumber
    p(\rvz) = \gN(\rvz \vert \bm{0}, \mI_{d \times d}).
\end{equation}
The posterior distribution $p(\rvz | \rvx)$ is unknown. However, we assume an approximation to it, which we denote by 
\begin{equation}
\label{eq:split_pz}    
      q(\rvz | \rvx) = \eta \gN(\rvz \vert \vmu_1, \mSigma_1) + (1-\eta) \gN(\rvz \vert \vmu_2, \mSigma_2),
\end{equation}
where $\vmu_1, \mSigma_1, \vmu_2, \mSigma_2$ depend on $\rvx$ and are generated by the encoder network and the dimension reduction component (explained below) the default choice for the mixture parameter  is $\eta = 5/6$  (the low sensitivity of our method to the choice of $\eta$ is demonstrated in 
the supplementary materials).  The first mode in \eqref{eq:split_pz} represents the inliers and the second one represents the outliers. 
We model $p(\rvz)$ as a single Gaussian since we only want to include the inlier information, while having the simplest possible design. In \S\ref{subsec:var}, we will numerically compare with the modeling of $p(\rvz)$ as a GMM. In  
the supplementary materials we intuitively clarify the mechanism that helps in such modeling.

The dimension reduction component involves a mapping from a higher-dimensional space onto the latent space. It is analogous to the RSR layer \citep{lai2020robust} that projects encoded points onto the latent space, but requires a more careful design since we consider a distribution rather than sample points. Due to this reduction, we assume that the mapped covariance matrices of $\rvz|\rvx$ are full, unlike common single-mode VAE models that assume a diagonal covariance \citep{kingma2014auto, an2015variational}.
We assume that the inliers lie on a low-dimensional structure and we thus enforce the lower rank $d/2$ for $\mSigma_1$, but allow $\mSigma_2$ to have full rank $d$. Nevertheless, we later describe a necessary regularization of both matrices. We remark that the low rank assumption results in the main distinction between the inliers and outliers in \eqref{eq:split_pz} (as noted in 
the supplementary materials the choice of $\eta>0.5$ is not crucial).

The unknown posterior distribution $p(\rvz | \rvx)$ is approximated within the variational family $\gQ = \{q(\rvz|\rvx)\}$ indexed by $\vmu_1$, $\mSigma_1$, $\vmu_2$ and $\mSigma_2$. 
Unlike a standard VAE, which maximizes the evidence lower bound (ELBO), MAW maximizes the following loss function, which
uses the $W_1$ distance (defined in the supplementary materials), instead of the KL divergence, for regularizing the log-likelihood of the data distribution: 
\begin{equation}
\label{eq:elbow}
    \mathcal{L}_\textup{MAW}(q) = \E_{p(\rvx)} \E_{q(\rvz|\rvx)} \log p(\rvx|\rvz) - W_1(q(\rvz), p(\rvz)) ~.
\end{equation}
We use the Wasserstein distance since it is more robust to outliers than the KL divergence and is thus more suitable for detecting anomalies (see related guarantees in \S\ref{sec:theory}). 
% Nevertheless, the use of the Wasserstein distance does not yield a lower bound on the log likelihood, unlike the KL divergence \citep{kingma2014auto}. However, the lower bound of the log likelihood is just an interpretative property of the traditional VAE and is not used in any theoretical guarantees.

Following the VAE framework, we use a Monte-Carlo approximation to estimate $\E_{q(\rvz|\rvx)} \log p(\rvx|\rvz)$ with 
i.i.d.~samples, $\{\rvz^{(t)}\}_{t=1}^T$, from $q(\rvz|\rvx)$ as follows:
\begin{equation}
\label{eq:monte_carlo}
    \E_{q(\rvz|\rvx)} \log p(\rvx|\rvz) \approx \frac{1}{T} \sum_{t=1}^T \log p(\rvx|\rvz^{(t)}).
\end{equation}
To enhance robustness, we let the negative log likelihood function $-\log p(\rvx|\rvz^{(t)})$ be proportional to the $\ell_{2}$ norm of the difference of the random variable $\rvx$ and a mapping of the sample $\rvz^{(t)}$ from $\R^d$ to $\R^D$ by the decoder, $\gD$, that is, 
\begin{equation}
\label{eq:logp_l1}
   -\log p(\rvx|\rvz^{(t)}) \propto \norm{\rvx - \gD(\rvz^{(t)})}_2 ~.
\end{equation}
We deviate from the common choice of the squared $\ell_2$ norm, which corresponds to an underlying Gaussian likelihood and assume instead a likelihood with a heavier tail.

MAW trains its networks by minimizing 
$-\mathcal{L}_\textup{MAW}(q)$. 
For $1 \leq i \leq L$,  
it samples $\{\rvz_{\textup{gen}}^{(i,t)}\}_{t=1}^T$ 
from $q(\rvz|\rvx^{(i)})$, where all samples
are independent. Using the aggregation formula $q(\rvz) = L^{-1} \sum_{i=1}^L q(\rvz|\rvx^{(i)})$, the approximation of 
$p(\rvx)$ by the empirical distribution of the training data, and \eqref{eq:elbow}-\eqref{eq:logp_l1},
MAW applies the following approximation of $-\mathcal{L}_\textup{MAW}(q)$:
\begin{equation}\label{eq:loss_in_general}
\begin{split}
    \frac{1}{LT} \sum_{i=1}^L  \sum_{t=1}^T &\norm{\rvx^{(i)} - \gD(\rvz_{\textup{gen}}^{(i,t)})}_2
         \\& +  W_1\left( \frac{1}{L} \sum_{i=1}^L q(\rvz|\rvx^{(i)}), p(\rvz)  \right).
\end{split}
\end{equation}
Our procedure of minimizing \eqref{eq:loss_in_general} is described in \S\ref{subsec:details}. 
% It is independent of the multiplicative constant in \eqref{eq:logp_l1} and therefore this constant is ignored in \eqref{eq:loss_in_general}.

During testing, MAW 
identifies outliers according to 
low similarity scores computed between test points and points generated from the learned inlier component of $\rvz|\rvx$.

\subsection{Details of Implementing MAW}\label{subsec:details}

MAW has a VAE-type structure with additional WGAN-type structure for minimizing the $W_1$ loss in \eqref{eq:loss_in_general}. 
We provide here details of implementing these structures. Some specific choices of the networks are described in \S\ref{sec:experiment} since they may depend on the type of datasets.

The VAE-type structure of MAW contains three ingredients: encoder, dimension reduction component and decoder. The encoder forms a neural network (NN), $\gE$, that maps the training sample $\rvx^{(i)}$ in $\R^{D}$ to $\vmu_{0, 1}^{(i)}, \vmu_{0, 2}^{(i)}, \vs_{0, 1}^{(i)}, \vs_{0, 2}^{(i)}$ in $\sR^{D'}$, where our default choice is $D' = 128$. 
The dimension reduction component then computes the following statistical quantities of the GMM $\rvz | \rvx^{(i)}$: means $\vmu_1^{(i)}$ and $\vmu_2^{(i)}$ in $\R^d$ and covariance matrices $\mSigma_1^{(i)}$ and $\mSigma_2^{(i)}$ in $\R^{d \times d}$.  First, a linear layer, represented by $\mA \in \sR^{D' \times d}$, maps (via $\mA^\T$) the features $\vmu_{0, 1}^{(i)}$, $\vmu_{0, 2}^{(i)} \in \R^{D'}$ to the following respective vectors in $\R^d$: 
$$\vmu_1^{(i)} = \mA^\T \vmu_{0, 1}^{(i)} ~\text{ and }~ \vmu_2^{(i)} = \mA^\T \vmu_{0, 2}^{(i)}.$$ 

The mapping of the covariance matrices is constructed as follows. Form $\mM_j^{(i)} = \mA^\T \textup{diag}(\vs_{0, j}^{(i)}) \mA$ for $j=1,2$. For $j=2$, compute $\mSigma_2^{(i)} = \mM_2^{(i)} \mM_2^{(i)\T}$. For $j=1$, we first need to reduce the rank of $\mM_1^{(i)}$. For this purpose, we form
\begin{equation}
\label{eq:decompose_M1}
\mM_1^{(i)} = \mU_1^{(i)} \textup{diag}(\bm{\sigma}_1^{(i)}) \mU_1^{(i)\T},
\end{equation}
the spectral decomposition of $\mM_1^{(i)}$, and then truncate its bottom $d/2$ eigenvalues. That is, let     $\tilde{\bm{\sigma}}_1^{(i)}$ $\in \R^d$ have the same entries as the largest $d/2$ entries of  $\mathbf{\bm{\sigma}}_1^{(i)}$ and zero entries otherwise. Then, compute
\begin{equation}
\label{eq:synthesis_M1}
    \tilde{\mM}_1^{(i)} = \mU_1^{(i)} \textup{diag}(\tilde{\bm{\sigma}}_1^{(i)}) \mU_1^{(i)\T} 
\end{equation}
and $$\mSigma_1^{(i)} = \tilde{\mM}_1^{(i)} \tilde{\mM}_1^{(i)\T}.$$
To ensure numerically-significant positive definiteness of both $\mSigma_1^{(i)}$ and $\mSigma_2^{(i)}$, we add to them an identity matrix. Despite this, the low-rank structure of $\mSigma_1^{(i)}$ is still evident. Note that the dimension reduction component only trains $\mA$. The decoder, $\gD: \R^{d} \to \R^{D}$, maps independent
samples, $\{\rvz_{\textup{gen}}^{(i,t)}\}_{t=1}^T$, generated for each $1 \leq i \leq L$ by the distribution 
    $$\eta \gN(\vmu_1^{(i)}, \mSigma_1^{(i)}) + (1-\eta) \gN(\vmu_2^{(i)}, \mSigma_2^{(i)}),$$ into the reconstructed data space.
    
The loss function associated with the VAE structure is the first term in \eqref{eq:loss_in_general}. We can write it as
\begin{equation}\label{eq:vae_loss}
        L_{\rm{VAE}}(\gE, \mA, \gD) = 
        {\frac{1}{LT}} \sum_{i=1}^L {\sum_{t=1}^T} \norm{\rvx^{(i)} -
        \gD(\rvz_{\textup{gen}}^{(i,t)})
         }_2~.
\end{equation} 
The dependence of this loss on $\gE$ and $\mA$ is implicit, but follows from the fact that the parameters of the sampling distribution of each  $\rvz_{\textup{gen}}^{(i,t)}$ were obtained by $\gE$ and $\mA$.

The WGAN-type structure seeks to minimize the second term in \eqref{eq:loss_in_general} using the dual formulation
\begin{align}\label{eq:dual_W1}
    &W_1 \left(\frac{1}{L} \sum_{i=1}^L q(\rvz|\rvx^{(i)}), p(\rvz) \right) =
    \\&\sup_{\substack{ \norm{f}_{\text{Lip}} \leq 1}} \E_{\rvz_{\textup{hyp}}\sim p(\rvz)}f(\rvz_{\textup{hyp}}) - \E_{\rvz_{\textup{gen}}\sim \frac{1}{L}\sum_{i=1}^L q(\rvz|\rvx^{(i)})}f(\rvz_{\textup{gen}}). \nonumber
\end{align}

The generator of this WGAN-type structure is composed of the encoder $\gE$ and the dimension reduction component, which we represent by $\mA$. It generates the samples 
$\{\rvz_{\textup{gen}}^{(i,t)}\}_{i=1,t=1}^{L,T}$ described above. The discriminator, $\Dis$, of the WGAN-type structure plays the role of the Lipschitz function $f$ in \eqref{eq:dual_W1}. It compares the latter samples with the i.i.d.~samples $\{\rvz_{\textup{hyp}}^{(i,t)}\}_{t=1}^{T}$ 
from the prior distribution. In order to make $\Dis$ 
Lipschitz, its weights are clipped to $[-1,1]$ during training. 
In the MinMax game of this WGAN-type structure, the discriminator minimizes and the generator ($\gE$ and $\mA$) maximizes
\begin{equation}
 \begin{split} \label{eq:wass_loss} 
        L_{W_{1}}(\Dis) =
         \frac{1}{LT} \sum_{i=1}^L \sum_{t=1}^T  \left(\Dis(\rvz_{\textup{gen}}^{(i,t)}) -  \Dis(\rvz_{\textup{hyp}}^{(i,t)})\right)~.
\end{split}   
\end{equation}
We note that maximization of \eqref{eq:wass_loss} by the generator is equivalent to minimization of the loss function 
\begin{equation}\label{eq:enc_loss}
  L_{\rm{GEN}}(\gE, \mA) 
   =   
   - \frac{1}{LT} \sum_{i=1}^L \sum_{t=1}^T \Dis(\rvz_{\textup{gen}}^{(i,t)})~.
\end{equation}
During training, MAW alternatively minimizes the losses \eqref{eq:vae_loss}, \eqref{eq:wass_loss} and \eqref{eq:enc_loss}
instead of their weighted sum.
Therefore, any multiplicative constant in front of either term of \eqref{eq:loss_in_general} will not affect the optimization. In particular, it was okay to omit the multiplicative constant of \eqref{eq:logp_l1} when deriving \eqref{eq:loss_in_general}.

% We remark that sliced Wasserstein distance \citep{kolouri2019generalized} and Sinkhorn algorithms \citep{cuturi2013sinkhorn} may efficiently calculate the Wasserstein distance. However, our choice of an auxiliary WGAN is a simple implementation to compute $W_1$ distance. For instance, learning the GMM using the sliced Wasserstein distance~\citep{kolouri2019generalized} relies on an EM-like approach, where the partial derivatives are very complicated. Moreover, the use of Sinkhorn algorithms is not straightforward, since they iteratively calculate the distance but do not minimize the distance with respect to the parameters within the neural network. 
% We noted the work on Sinkhorn autoencoders~\citep{patrini2020sinkhorn}, which is related, but does not directly apply to our methodology.

For each testing point $\rvy^{(j)}$, we sample $\{\rvz_{\textup{in}}^{(j,t)}\}_{t=1}^T$ from the inlier mode of the learned latent Gaussian mixture and decode them as $\{\tilde{\rvy}^{(j,t)}\}_{t=1}^T =\{\gD(\rvz_{\textup{in}}^{(j,t)})\}_{t=1}^T$. Using a similarity measure $S(\cdot,\cdot)$ (our default is the cosine similarity), we compute
    $$S^{(j)} = %\frac{1}{T} 
    \sum_{t=1}^T S(\rvy^{(j)}, \tilde{\rvy}^{(j,t)}).$$
If $S^{(j)}$ is larger than a chosen threshold, 
then $\rvy^{(j)}$ is classified as normal, and otherwise, novel. Additional details of MAW are in 
the supplementary materials.

We remark that in our setting we find it natural to implement an auxiliary WGAN on top of the VAE component in order to estimate the $W_1$ distance. We did not find it useful to directly estimate the $W_1$ distance by either the sliced Wasserstein distance \citep{kolouri2018sliced, kolouri2019generalized} or the Sinkhorn algorithm \citep{cuturi2013sinkhorn}. Indeed, it is not clear how to use these methods in order to minimize the estimated $W_1$ distance with respect to the parameters within the neural network. 
In particular, the partial derivatives for learning the GMM using the sliced Wasserstein distance already have very complicated forms, and it is very difficult to include them in our framework, where neural networks are involved.

\section{THEORETICAL GUARANTEES}\label{sec:theory}

We theoretically establish the superiority of using the Wasserstein distance over the KL divergence, where we leave out some details (in particular proofs) to  the supplementary materials. 
We formulate a mathematical setting that aims to isolate the minimization of the WGAN-type structure introduced in \S\ref{subsec:details},
while ignoring unnecessary complex components of MAW. We assume a mixture parameter $\eta>1/2$, a separation parameter  $\epsilon >0$ and denote by $\cal R$ the regularizing function, which can be either the KL divergence or the Wasserstein distance, and by ${\cal S}_{+}^{K}$ and ${\cal S}_{++}^{K}$ the sets of $K \times K$ positive semidefinite and positive definite matrices, respectively. 
Our mathematical setting, which we motivate in  
the supplementary materials, assumes $\bm{\mu}_0  \in \R^K$ and $\mSigma_0 \in {\cal S}_{++}^{K}$ and requires to minimize
\begin{align}\label{eq:bary}
    \min_{\substack{\vmu_1, \vmu_2 \in \R^K; \mSigma_1, \mSigma_2 \in {\cal S}_{+}^{K}   \\ \rm{s.t.}~ \norm{\vmu_1 - \vmu_2 }_2 \geq \epsilon}}
 &\eta {\cal R} \left( \gN(\vmu_1, \mSigma_1),  \gN(\vmu_0, \mSigma_0) \right) 
 \\&+ (1-\eta) {\cal R} \left( \gN(\vmu_2, \mSigma_2),  \gN(\bm{\mu_0}, \mSigma_0) \right).\nonumber
\end{align}

This minimization aims to approximate the ``prior'' distribution $\gN(\vmu_0, \mSigma_0)$ with a Gaussian mixture distribution. For MAW, $\vmu_0 = \bm{0}$ and $\mSigma_0 = \mI$, but our generalization helps clarify things. The constraint $\norm{\vmu_1 - \vmu_2}_2 \geq \epsilon$ distinguishes between the inlier and outlier modes and it is a realistic assumption as long as $\epsilon$ is sufficiently small. 

\subsection{Guarantees for \eqref{eq:bary} with Identical Covariances}\label{subsuec:id_cov}

Our cleanest result is when $\mSigma_0$, $\mSigma_1$ and $\mSigma_2$ coincide. It is formulated next and demonstrates robustness to the outlier component by the $W_1$ (or $W_p$, $p \geq 1$) minimization and not by the KL minimization (its proof is in the supplementary materials).
\begin{proposition}\label{prop:baryW1andKL} 
If $\vmu_0 \in \R^K$, $\mSigma_0  \in {\cal S}_{++}^{K}$, $\epsilon > 0$ and $1>\eta>1/2$, then the minimizer of \eqref{eq:bary} with ${\cal R} = W_p$, $p \geq 1$ and the additional constraint: $\mSigma_0 = \mSigma_1 = \mSigma_2$, satisfies $\vmu_1=\vmu_0$, and thus the recovered inlier distribution coincides with the ``prior distribution''. 
However, the minimizer of \eqref{eq:bary} with ${\cal R} = KL$ and the same constraint satisfies $\vmu_0 = \eta \vmu_1 + (1-\eta) \vmu_2$.
\end{proposition}

That is, under the above setting with ${\cal R} = W_1$, the estimated mean of the inlier distribution, $\mu_1$, coincides with the mean of the prior distribution, independently of the outlier distribution. However, when ${\cal R} = KL$, the estimated mean of the inlier distribution is sensitive to outliers.

%%%%%%%%%%%%%%%%%%%%%%%%%%%%%%%%%%%%%%%%%%%%%%%%%%%%%%%%%%%%%%%%%%%%%%%%%%%
\subsection{Guarantees for \eqref{eq:bary} with Low-rank \boldmath{$\Sigma_1$}}\label{subsuec:low_cov}
We study the minimization problem \eqref{eq:bary} when $\mSigma_1$ has a low rank and $\mSigma_2 \in {\cal S}_{++}^K$. 
We fully analyze the cases where ${\cal R} = W_2$ and ${\cal R} = KL$; however, the case where  ${\cal R} =  W_1$ is difficult to analyze and compute. 
We first formulate results for both cases (${\cal R} =  W_2$ and ${\cal R} =  KL$), and then clarify them. When ${\cal R} = W_2$, we assume that the prior distribution has  zero mean vector $\vmu_0 = \bm{0}_K \in {\mathbb R}^K$ and covariance $\mSigma_0 = \bm{I}_{K\times K} \in {\mathbb R}^{K \times K}$. We further denote by $\bm{1}_K$ the vector $(1, \cdots, 1) \in \R^K$. 
Similarly, we denote for any $n \in \mathbb{N}$, $\bm{0}_n$, $\bm{1}_n$, $\mI_{n \times n}$.
For vectors $\va \in \R^{n}$ and $\vb \in \R^m$, we denote the concatenated vector in $\R^{n+m}$ by $(\va; \vb)$.

\begin{proposition}\label{prop:baryW2}
If $\kappa$, $K \in \mathbb{N}$, $K>\kappa \geq1$, $\epsilon>0$, $1>\eta> \eta^{\star} :=\frac{K-\kappa+\epsilon^2}{K-\kappa+2\epsilon^2}$, 
$u^{\star} := \left( \frac{(K-\kappa)(1-\eta)}{\epsilon^2 (2\eta-1)} \right)^{\frac{1}{3}}$, where one can note that $\eta^{\star} > \frac{1}{2}$ and $ u^{\star} \in (0,1)$, then the minimizer of \eqref{eq:bary} with ${\cal R} = W_2$ and the constraints that $\mSigma_1$ is of rank $\kappa$ and $\mSigma_2$ is of rank $K$, 
satisfies $\bm{0}_K = u^{\star}\bm{\mu_2} + (1-u^{\star})\bm{\mu_1} $, $\mSigma_1=\textup{diag}(\bm{1}_\kappa; \bm{0}_{K-\kappa})$ and $\mSigma_2=\textup{diag}(\bm{1}_\kappa; {(u^{\star})}^{-2}\bm{1}_{K-\kappa})$. Moreover, $\norm{\vmu_1}_2 = u^{\star} \epsilon$ and $\norm{\vmu_2}_2 = (1-u^{\star}) \epsilon$.
\end{proposition}

\begin{proposition}\label{prop:baryKL}
If $\kappa$, $K \in \mathbb{N}$, $K>\kappa \geq1$, $\epsilon>0$, $\eta>0$, $\vmu_0$, $\vmu_1 \in \R^K$, $\mSigma_0 \in {\cal S}_{++}^{K}$ and $\mSigma_1 \in {\cal S}_{+}^{K}$, \textup{rank}$( \mSigma_1)= \kappa$, then 
\begin{equation*}
    KL(\mathcal{N}(\vmu_1, \mSigma_1) || \mathcal{N}(\vmu_0, \mSigma_0)) = \infty.
\end{equation*}
Thus, the solution of \eqref{eq:bary} with ${\cal R} = KL$ and the additional constraints $\mathrm{rank}(\mSigma_1)=\kappa$ and $\mSigma_0 = \bm{I}$ is ill-posed.  
\end{proposition}

Note that Proposition~\ref{prop:baryW2} implies that as $\eta \rightarrow 1$, $u^{\star} \rightarrow 0$. Hence for the inlier component 
$\vmu_1 \rightarrow \bm{0}_K$ as $\eta \rightarrow 1$ and $\mSigma_1=\textup{diag}(\bm{1}_\kappa; \bm{0}_{K-\kappa})$. Therefore, in the limit, the inlier distribution has the same mean as the prior distribution. Furthermore, its covariance is obtained by an appropriate projection of the covariance $\mSigma_0$ onto a $\kappa$-dimensional subspace,  independently of $\eta$. 
We similarly note that as $\eta \rightarrow 1$, $\bm{\Sigma}_2 \rightarrow \textup{diag}(\bm{1}_\kappa; \boldsymbol{\infty}_{K-k})$, so that the outliers disperse. 
The supplementary materials include the proof of Proposition \ref{prop:baryW2} and a discussion that clarifies why the formulation and proof of Proposition \ref{prop:baryW2} are not sufficient for inferring the effect of the $W_1$ minimization on MAW. 

Proposition~\ref{prop:baryKL} implies that the KL divergence is unsuitable for low-rank covariance modeling as it leads to an infinite value in the optimization problem.

%%%%%%%%%%%%%%%%%%%%%%%%%%%%%%%%%%%%%%%%%%%%%%%%%%%%%%%%%%%%%%%%%%%%%%%%%%%

\section{EXPERIMENTS}\label{sec:experiment}
We describe the competing methods and experimental choices in \S\ref{sec:choice_MAW}. We report on the comparison with the competing methods in \S\ref{subsec:res}. We demonstrate the importance of the novel features of MAW in \S\ref{subsec:var}.

\subsection{Competing Methods and Experimental Choices} \label{sec:choice_MAW}

We compared MAW with the following %state-of-the-art 
methods (descriptions and code links are in   
the supplementary materials): Deep Autoencoding Gaussian Mixture Model (DAGMM) \citep{zong2018deep}, Deep Structured Energy-Based Models (DSEBMs) \citep{zhai2016deep}, Isolation Forest (IF) \citep{liu2008isolation},  
Local Outlier Factor (LOF) \citep{breunig2000lof}, One-class Novelty Detection Using GANs (OCGAN) \citep{perera2019ocgan}, One-Class SVM (OCSVM)  \citep{heller2003one} and %Robust Subspace Recovery
RSR Autoencoder (RSRAE) \citep{lai2020robust}.

We remark that IF, LOF and RSRAE were originally proposed for outlier detection and we thus apply their trained model for detecting novelties in 
the test data.

For MAW and the above four reconstruction-based methods, that is, DAGMM, DSEBMs, OCGAN and RSRAE, we use the following structure of encoders and decoders, which vary with the type of data (images or non-images). 
For non-images, which are mapped to feature vectors of dimension $D$, the encoder is a fully connected network with output channels $(32, 64, 128, 128 \times 4)$. The decoder is a fully connected network with output channels $(128, 64, 32, D)$, followed by a normalization layer at the end. 
For image datasets, the encoder has three convolutional layers with output channels $(32, 64, 128)$, kernel sizes $(5\times5, 5\times5, 3\times3)$ and strides $(2, 2, 2)$. Its output is flattened to lie in $\R^{128}$ and then mapped into a $128 \times 4$ dimensional vector using a dense layer (with output channels $128 \times 4$). 
The decoder of image datasets first applies a dense layer from $\R^2$ to $\R^{128}$ and then three deconvolutional layers with output channels $(64, 32, 3)$, kernel sizes $(3\times3, 5\times5, 5\times5)$ and strides $(2, 2, 2)$.
For all experiments, the MAW discriminator is a fully connected network with size $(32, 64, 128, 1)$.

For MAW we set the following parameters, where additional  details are in 
the supplementary materials. 
Intrinsic dimension: $d=2$; mixture parameter: $\eta = 5/6$,  sampling number: $T=5$, and size of $\mA$ (used for dimension reduction): $128 \times 2$. We further test the sensitivity of MAW to  changes of the hyperparameters $d$ and $\eta$ in  
the supplementary materials. The code is available at \url{https://github.com/JCL823/MAW}.

\subsection{Comparison of MAW with State-of-the-art Methods}\label{subsec:res}

\begin{figure}[ht!]
\centering
\includegraphics[width=\columnwidth]{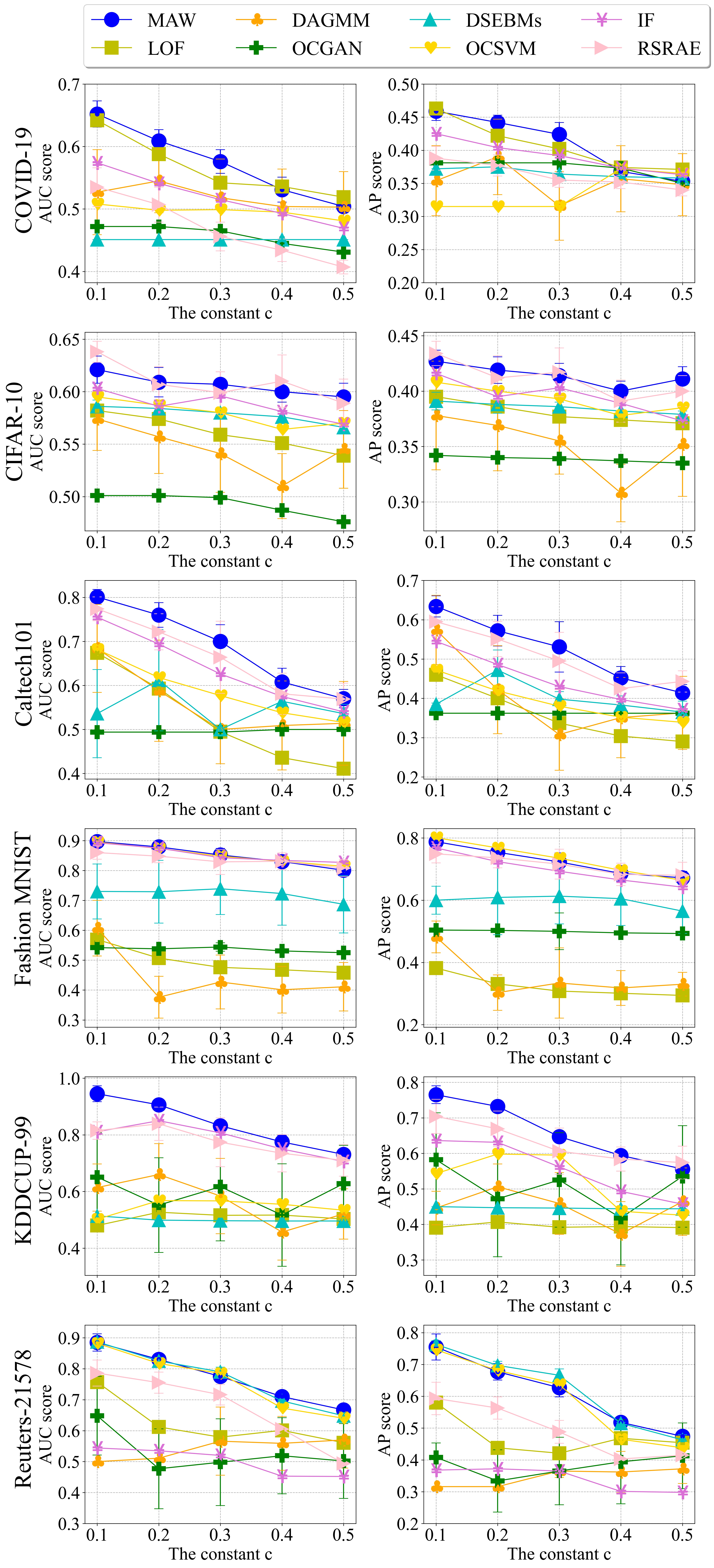}
\caption{AUC (on left) and AP (on right) scores with training ratio of outliers per inliers $c = 0.1$, $0.2$, $0.3$, $0.4$ and $0.5$ for the six datasets.} 
\label{fig:res}
\end{figure}

% For each dataset, we arbitrarily fix a class and uniformly sample $N$ training inliers and $N_{\textup{test}}$ testing inliers from that class. We let $N=160$, $450$, $100$, $300$, $6000$, $350$ and $N_{\textup{test}} =60$, $150$, $100$, $60$, $1200$, $140$  for COVID-19, CIFAR-10, Caltech101, Fashion MNIST, KDDCUP-99 and Reuters-21578, respectively. We fix $c$ in $\{ 0.1, 0.2, 0.3, 0.4, 0.5 \}$, and uniformly sample outliers for training from the rest of the clusters, while maintaining a fraction of $c$ outliers per inliers.
% We fix $c_{\textup{test}}$ 
% in $\{ 0.1, 0.3, 0.5, 0.7, 0.9 \}$
% and uniformly sample outliers from the rest of the clusters for testing, while maintaining a fraction of $c_{\textup{test}}$ per inliers. 
% Table~\ref{tab:numdata} lists for each dataset the data types, numbers of clusters, dimensions, numbers of instances and numbers of inliers and outliers.

We use six datasets for novelty detection: COVID-19 Radiography database~\citep{chowdhury2020can}, CIFAR-10~\citep{krizhevsky2009learning}, Caltech101~\citep{fei2004learning}, Fashion MNIST~\citep{xiao2017fashion}, KDDCUP-99~\citep{dua2017uci} and Reuters-21578~\citep{lewis1997reuters}. We distinguish between image datasets (COVID-19, CIFAR-10, Catlech101 and Fashion MNIST) and non-image datasets (KDDCUP-99 and Reuters-21578).
We describe each dataset, common preprocessing procedures and choices of their largest clusters in  
the supplementary materials. Each dataset contains several clusters (3 for COVID-19, 10 for CIFAR-10, 11 largest ones for Caltech101, 10 for Fashion MNIST, 2 for KDDCUP-99 and 5 largest ones for Reuters-21578, respectively). We arbitrarily fix a class and uniformly sample $N$ training inliers and $N_{\textup{test}}$ testing inliers from that class. We let $N=$ $160$, $450$, $100$ , $300$, $6000$, $350$ and $N_{\textup{test}} =$  $60$, $150$, $100$ , $60$, $1200$, $140$  for COVID-19, CIFAR-10, Caltech101, Fashion MNIST, KDDCUP-99 and Reuters-21578, respectively. We fix $c$ in $\{ 0.1, 0.2, 0.3, 0.4, 0.5 \}$, and uniformly sample outliers for training from the rest of the clusters, while maintaining a fraction of $c$ outliers per inliers.
We also fix $c_{\textup{test}}$ 
in $\{ 0.1, 0.3, 0.5, 0.7, 0.9 \}$
and uniformly sample outliers from the rest of the clusters for testing, while maintaining a fraction of $c_{\textup{test}}$ per inliers.

Using all possible thresholds for the finite datasets, we compute the AUC (area under curve) and AP (average precision) scores, while considering the outliers as “positive”. For each fixed $c = 0.1$, $0.2$, $0.3$, $0.4$, $0.5$ we average these results over the values of $c_{\textup{test}}$, the different choices of an inlier cluster (among all possible clusters), and three runs with different random initializations for each of these choices. We also compute the corresponding standard deviations. We report these results in Fig.~\ref{fig:res} and further specify numerical values in  
the supplementary materials. We observe state-of-the-art performance of MAW in all of these datasets. There are very special instances, where other methods perform better, for example, in Reuters-21578, DSEBMs performs slightly better than MAW and OCSVM has comparable performance. However, overall MAW is the most competitive method considering all instances. 
In the supplementary materials we compare the runtime of MAW with benchmark methods and further study the accuracy of MAW in a different scenario, where the outliers of the training and test sets have different characteristics. We show that in this scenario MAW performs even better than the regular scenario.

\subsection{Testing the Effect of the Novel Features of MAW}\label{subsec:var}

We experimentally validate the effect of the following  features of MAW: the least absolute deviation for reconstruction, the $W_1$ metric for the regularization of the latent distribution, the GMM assumption, full covariance matrices resulting from the dimension reduction component, the lower rank constraint for the inlier mode and the use of a single mode prior distribution. To this end, we consider the following alternative models.

\noindent
\textbf{MAW-MSE:}  It replaces the least absolute deviation loss $L_{\rm{VAE}}$
     with the common mean squared error (MSE). 
    %  That is, it replaces $\norm{\rvx^{(i)} -
    %     \gD(\rvz_{\textup{gen}}^{(i,t)})
    %      }_2$ in \eqref{eq:vae_loss} with $\norm{\rvx^{(i)} -
    %     \gD(\rvz_{\textup{gen}}^{(i,t)})
    %      }_2^2$.
     
\noindent
\textbf{MAW-KL divergence:} It replaces the Wasserstein distance in \eqref{eq:dual_W1} with the KL-divergence. 
% Thus, it replaces the WGAN-type structure of the discriminator with a standard GAN.
      
\noindent
\textbf{MAW-same rank:} It uses the same rank $d$ for both $\mSigma_1^{(i)}$ and $\mSigma_2^{(i)}$, instead of forcing $\mSigma_1^{(i)}$ to have lower rank $d/2$.
     
\noindent
\textbf{MAW-single Gaussian:} It replaces the GMM for the latent distribution with a single Gaussian with a full covariance matrix.
      
\noindent
\textbf{MAW-diagonal cov.:} It replaces the full covariance matrices resulting from the dimension reduction component by diagonal covariances. Its encoder directly produces 2-dimensional means and diagonal covariances (one of rank 1 for the inlier mode and one of rank 2 for the outlier mode). 
     %\item 
 
\noindent
\textbf{GMM prior:} It replaces the single standard normal distribution prior with a bi-modal Gaussian distribution. One mode is a standard normal distribution in $\mathbb{R}^d$ and the other is Gaussian with zero mean  and diagonal covariance matrix whose first $d/2$ diagonal elements are ones and the rest are zeros.

\noindent
\textbf{VAE:} It has the same encoder and decoder structures as MAW. Instead of a dimension reduction component, it uses a dense layer
 which maps the output of the encoder to a 4-dimensional vector composed of a 2-dimensional mean and 2-dimensional diagonal covariance. This is common for a traditional VAE.

\begin{figure}[h!]
\centering
\includegraphics[width=\columnwidth]{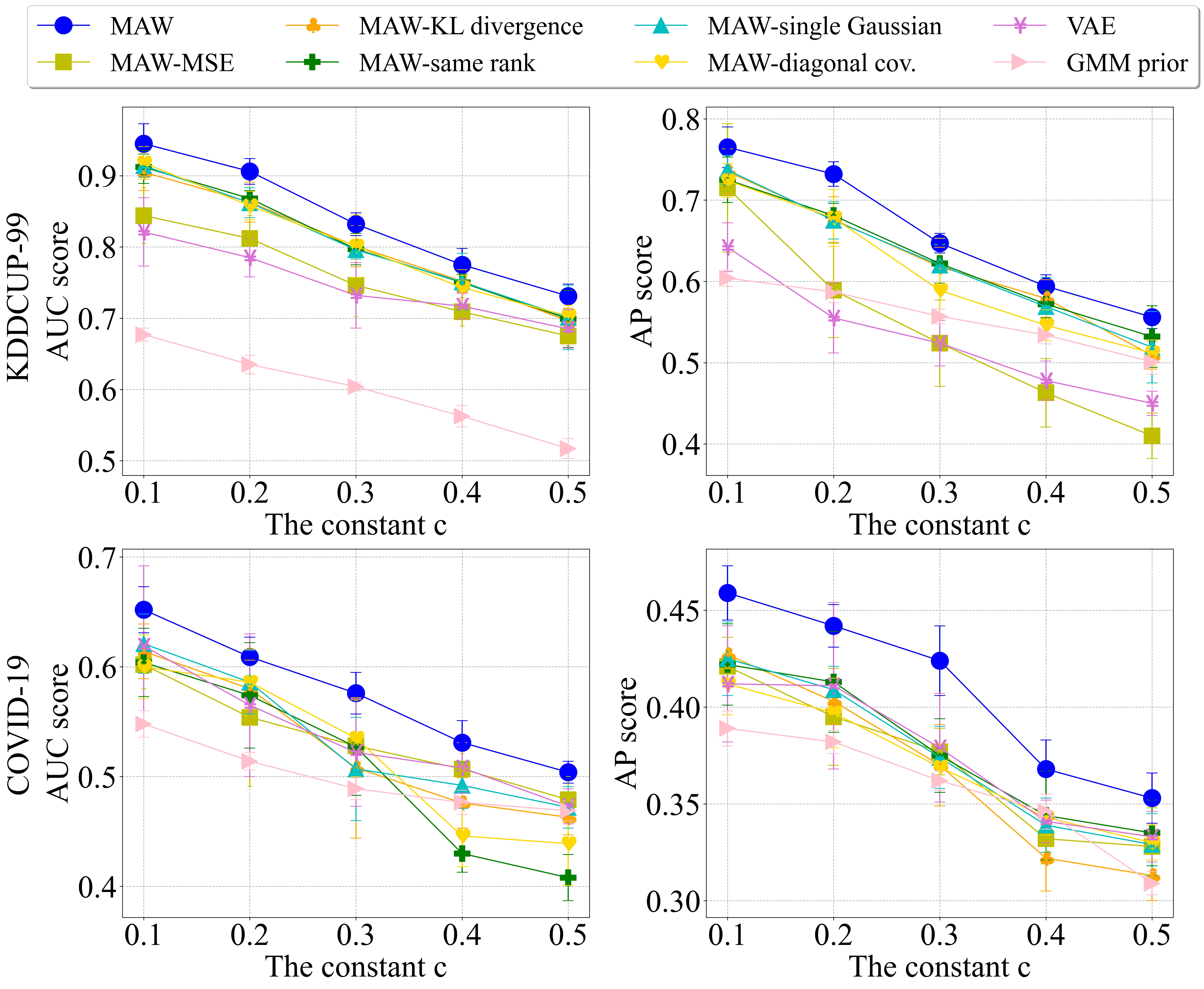}
\caption{AUC (on left) and AP (on right) scores for variants of MAW (missing a novel component) with training ratio of outliers per inliers $c = 0.1$, $0.2$, $0.3$, $0.4$ and $0.5$, using KDDCUP-99 and COVID-19.}
\label{fig:variations}
\end{figure}

We compared the above 7 methods with MAW using two datasets: KDDCUP-99 and COVID-19 with training ratio of outliers per inliers $c = 0.1$, $0.2$, $0.3$ , $0.4$ and $0.5$. We followed the experimental setting described in \S \ref{sec:choice_MAW}. Fig.~\ref{fig:variations} reports the averages and standard deviations of the computed AUC and AP scores, where the corresponding numerical values are further recorded in 
the supplementary materials. 
The results indicate a clear decrease of accuracy when missing any of the novel components of MAW or using a standard VAE (i.e., ``VAE''). 
Nevertheless, the use of a single diagonal matrix in ``VAE'' can help decrease the capacity of the latent distribution and thus ``VAE'' may perform better than the variants of MAW (but not MAW).  
In some cases, the variants of MAW show a rather poor performance and we believe it is due to the following reasons:  
modeling the prior as a Gaussian mixture in ``GMM prior'' does not help with outlier detection; the use of the full covariance in ``MAW-single Gaussian'' may result in high capacity; 
``MAW-MSE'', ``MAW-KL divergence'' and ``MAW-same rank'' do not ensure either robustness or low-rank modeling for the inliers, and thus may significantly increase the capacity of the model so that learning from outliers is easier (especially for large $c$); and ``MAW-diagonal cov.'' may limit the covariance of the outliers (though it is often at least comparable to VAE).

\subsection{Further Validation of GMM}\label{subsec:further_support_gmm}

\begin{figure}[t!]
\centering
\includegraphics[width=\columnwidth]{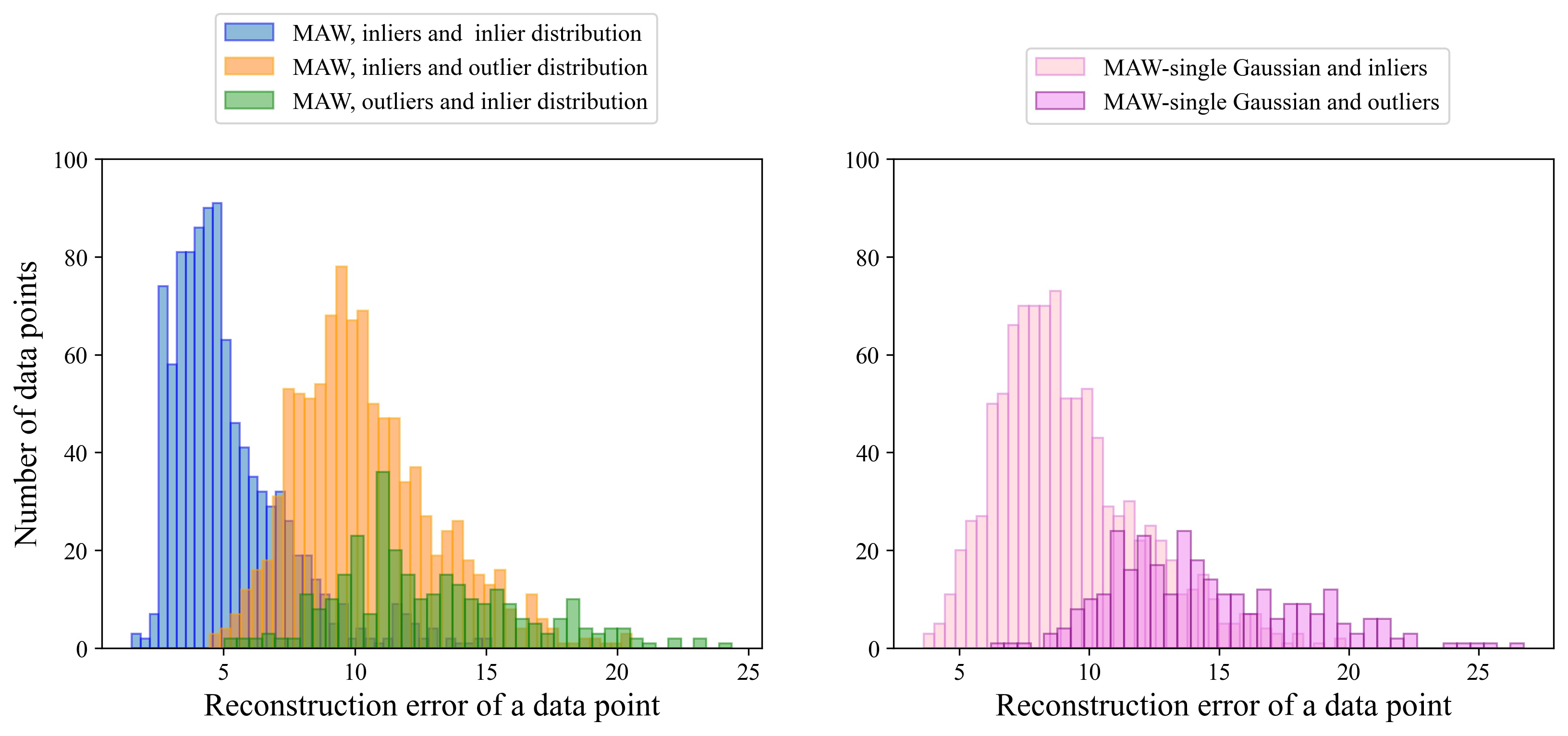}
\caption{Demonstration of the distributions of the three types of reconstruction errors obtained with MAW (left) and the two types obtained with MAW-single Gaussian (right).} 
\label{fig:recon_error}
\end{figure}

To further support our claim that the GMM is helpful for separating inliers and outliers in the latent space, we investigate the reconstruction errors of both MAW and 
MAW-single Gaussian of \S\ref{subsec:var} (which replaces the GMM with a single Gaussian distribution with a full rank). 
We use the KDDCUP-99 dataset with 1,000 inliers and 300 outliers in the training set, where the initial training of MAW (or MAW-single Gaussian) is the same as in \S\ref{sec:experiment}. In Fig.~\ref{fig:recon_error}, we demonstrate the reconstruction error distribution of data points according to the following five scenarios.
\begin{enumerate}%[leftmargin=12pt, parsep=0pt]
    \item \label{sce:1}\textbf{MAW, inliers and  inlier distribution:} 
    Apply the trained MAW (with the corrupted model) to the inliers of the training set, while using only the inlier mode in the latent code and compute the reconstruction error between the output and the input (the $\ell_2$ norm of their difference).   
    \item \label{sce:2}\textbf{MAW, inliers and outlier distribution:} Same  as case 1, but replace the inlier mode with the outlier mode. 
    \item \label{sce:3}\textbf{MAW, outliers and inlier distribution:} 
    Same as case 1, but replace the inliers (input of MAW) with the outliers.
    \item \label{sce:4}\textbf{MAW-single Gaussian and inliers:} Same as case 1, but replace MAW with MAW-single Gaussian.
    \item \label{sce:5}\textbf{MAW-single Gaussian and outliers:} 
    Same as 1, but replace the inliers (as input of the trained MAW-single Gaussian) with the outliers. 
\end{enumerate}

We can see from cases~\ref{sce:1} and~\ref{sce:2} above (which appear on the left of Fig.~\ref{fig:recon_error}) that if we try to reconstruct the inliers, then the reconstruction errors with the outlier mode are higher than those with the inlier mode. In particular, it is obvious that the inlier and outlier modes are different and do not collapse. Although we did not supervisedly train the inlier and outlier modes, it seems that the inliers align well with the inlier distribution. 
Moreover, comparing cases~\ref{sce:1} and~\ref{sce:3} above (still left of Fig.~\ref{fig:recon_error}), we can nicely distinguish between the distributions of the reconstruction errors of the inliers and the outliers. 
On the other hand, cases~\ref{sce:4} and~\ref{sce:5} (on the right of Fig.~\ref{fig:recon_error}) indicate that when using MAW-single Gaussian instead of MAW, the distributions of reconstruction errors of the inliers and outliers are indistinguishable. This experiment thus demonstrates the effectiveness of the GMM of MAW in separating the inliers and outliers for this particular experiment.

\section{CONCLUSION AND FUTURE WORK}\label{sec:conclusion}
We introduced MAW, a robust VAE-type framework for novelty detection that can tolerate high corruption of the training data. 
We proved that the Wasserstein distance used in MAW has better robustness to outliers and is more suitable to a low-dimensional inlier component than the KL divergence.
We demonstrated state-of-the-art performance of MAW with a variety of datasets and experimentally validated that omitting any of the new ideas results in a significant decrease of accuracy. 

We would like to indicate three limitations of MAW. First, there are some special instances, where other methods performed better than MAW, though overall MAW outperformed the rest of the methods. 
Second, MAW is slow. We expect that better implementation of its dimension reduction component can speed it up, so that it is as fast as other methods that use multiple neural networks. At last, MAW assumes the existence of both inlier and outlier modes for training (see assumptions of Props.~\ref{prop:baryW1andKL} and \ref{prop:baryW2}). Indeed, one may check that the performance of MAW (and RSRAE) are not as competitive when $c=0$. Since we assumed that the underlying distribution represented both inliers and outliers, we did not report such results.

MAW has practical applications of societal impact, such as  medical diagnosis.  
One potential negative impact can arise if MAW identifies outliers  due to their belonging to underrepresented groups.
In the future, we would thus like to explore the overall fairness of MAW, possible fairer versions of it and the tradeoff between robustness and fairness in our theoretical setting.

Another future plan is to extend and test some of our ideas for the problem of robust generation, in particular, for building generative networks which are robust against adversarial training data. 
We also hope to further extend our theoretical guarantees. For example, two problems that currently seem intractable are the study of the $W_1$ version of Proposition~\ref{prop:baryW2} and of the minimizer in \eqref{eq:minmixlatent} 
(which is a weaker version of \eqref{eq:bary}). 

\subsubsection*{Acknowledgements}

This work was partially supported by NSF award DMS 2124913 and the Kunshan Municipal Government
research funding.

% \clearpage
% \newpage
\bibliography{MAW_ref}

\appendix
\onecolumn

\begin{center}
    \Large{\textbf{SUPPLEMENTARY MATERIALS}}
\end{center}
We include additional explanations, proofs, demonstrations and experiments as follows:
\S\ref{sec:alg} further clarifies MAW and its implementation;  \S\ref{sec:sensitivity} examines the sensitivity of MAW to hyperparameters; \S\ref{sec:runtime} compares the runtime of MAW with benchmark methods; \S\ref{sec:mix_data} extends the previous numerical studies to a different type of outliers; \S\ref{sec:moretheory} extends our theoretical discussion and proves all the stated propositions; \S\ref{sec:benchmark} reviews the details of the benchmark methods;  \S\ref{sec:datasets} reviews the details of the datasets; and \S\ref{sec:numerical} provides numerical tables for the results plotted in the different figures.

\section{ADDITIONAL EXPLANATIONS AND IMPLEMENTATION DETAILS OF MAW}\label{sec:alg}

In \S\ref{sec:ELBO} we review the ELBO function and explain our robust version of ELBO.
% ELBOW is obtained from ELBO. 
The basic mechanism of MAW is clarified in \S\ref{subsec:mechanism}.  
Additional implementation details of MAW are in \S\ref{sec:MAW_implement}. At last, \S\ref{sec:alg_box} provides algorithmic boxes for training MAW and applying it for novelty detection.

% This section is organized as follows: \S\ref{sec:Wp} reviews the definition of the $p$-Wasserstein distance; \S\ref{sec:ELBO} 
% clarifies the direct relationship between $\mathcal{L}_\textup{MAW}$ and ELBO; \S\ref{subsec:proof}-\S\ref{subsec:proofbaryKL}
% prove Propositions~\ref{prop:baryW1andKL}--\ref{prop:baryKL}, respectively; \S\ref{subsec:W2theory} explains why Proposition~\ref{prop:baryW2} and its proof are not sufficient for explaining the effect of the $W_1$ optimization on MAW; finally, \S\ref{sec:numerical} provide table representations for Figs.~\ref{fig:res} and~\ref{fig:variations}.

% \subsection{Review of the $p$-Wasserstein Distance}
% \label{sec:Wp}

% For $p \geq 1$, we denote by $W_p$ the $p$-Wasserstein distance in $\R^D$. For two probability distributions, $\mu, \nu$ on $\R^D$,
% \begin{equation*}
% W_p(\mu, \nu) = \left( \inf_{\pi \in \Pi(\mu, \nu)} \E_{(\rvx,\rvy) \sim \pi} \norm{\rvx-\rvy}_2^p \right)^{1/p} ~,
% \end{equation*}
% where $\Pi(\mu, \nu)$ is the set of joint distributions with $\mu$ and $\nu$ as marginals.

\subsection{Obtaining $\mathcal{L}_\textup{MAW}$ by Modifying ELBO}
\label{sec:ELBO}

A standard VAE framework would minimize the expected KL-divergence from $p(\rvz|\rvx)$ to $q(\rvz|\rvx)$ in $\gQ$, where the expectation is taken over $p(\rvx)$. By Bayes' rule this is equivalent to maximizing the evidence lower bound (ELBO):
\begin{equation*}
    \textup{ELBO}(q) = \E_{p(\rvx)} \E_{q(\rvz|\rvx)} \log p(\rvx|\rvz) - \E_{p(\rvx)} KL(q(\rvz|\rvx) \Vert p(\rvz)) ~.
\end{equation*}
The first term of ELBO is the reconstruction likelihood. Its second term restricts the deviation of $q(\rvz|\rvx)$ from $p(\rvz)$ and can be viewed as a regularization term. 
$\mathcal{L}_\textup{MAW}$ is a more robust version of ELBO with a different regularization. 
Recall that for $p \geq 1$, we denote by $W_p$ the $p$-Wasserstein distance in $\R^D$. For two probability distributions, $\mu, \nu$ on $\R^D$,
\begin{equation*}
W_p(\mu, \nu) = \left( \inf_{\pi \in \Pi(\mu, \nu)} \E_{(\rvx,\rvy) \sim \pi} \norm{\rvx-\rvy}_2^p \right)^{1/p} ~,
\end{equation*}
where $\Pi(\mu, \nu)$ is the set of joint distributions with $\mu$ and $\nu$ as marginals.
MAW replaces 
$\E_{p(\rvx)} KL(q(\rvz|\rvx) \Vert p(\rvz))$ with $W_1(q(\rvz), p(\rvz))$. 
We remark that the $W_1$ distance cannot be computed between $q(\rvz|\rvx)$ and $p(\rvz)$ and $\mathcal{L}_\textup{MAW}$ thus practically replaces $q(\rvz | \rvx)$ with its expected distribution, $q(\rvz) = \E_{p(\rvx)} q(\rvz | \rvx)$ (or a discrete approximation of this). 

We emphasize that $\mathcal{L}_\textup{MAW}$ is not necessarily a lower bound of the likelihood. The $W_1$ distance in $\mathcal{L}_\textup{MAW}$ can rather be understood as a regularization involving the estimated posterior and prior distribution.

\subsection{Insights on the Mechanism of MAW}\label{subsec:mechanism}

We explain the basic mechanism of MAW for unsupervised alignment of the inliers with the inlier mode of the latent distribution. Since we do not have labels for the training set, we cannot supervisedly determine the inlier mode. Nevertheless, the robust losses (the least absolute deviation and the $W_1$ distance) guide the estimation of the inlier mode as they help in ignoring the effect of the outliers.
Least absolute deviation metrics have been shown to be robust to outliers in special mathematical settings \citep{lopuhaa1991,lerman2018overview, lai2020robust}. The robustness of the Wasserstein distance within a mathematical setting was studied in \S\ref{sec:theory} of the main text. Here we would like to provide some intuition on how the complex procedure of MAW succeeds by using these robust metrics.

Assume that the inliers are sampled from a distribution on a low-dimensional manifold that can be encoded by a Gaussian on a low-dimensional latent space. Assume further that the outliers are arbitrary, but their percentage is smaller. Given these assumptions, MAW aims to model the mixture component of the inliers in the latent space as a Gaussian with low-rank covariance (and that of the outliers as a Gaussian with full-rank covariance). 

\newpage

In order to provide some technical intuition for this model and show that it can fit the assumed data, let us suppose on the contrary that during training, inliers and outliers are assigned to the wrong modes, and show that this can either not happen or will be corrected.

We first assume a case of collapse during training, where both the inliers and outliers are modeled (in the latent space) by a Gaussian distribution with a low-rank covariance. In this case, the $W_1$ distance is minimized over a smaller set (due to the constraint on the rank of the outlier mode) and thus the loss is increased. 

We next assume another case of collapse during training, where both the inliers and outliers are modeled (in the latent space) by a full-rank Gaussian. In this case it is most likely that the minimizer for the inliers will be full-rank, and thus due to the assumed low-dimensional structure of the inliers, it will result in an increase of the reconstruction error. 

At last, assume that during training the inliers are modeled (in the latent space) by a Gaussian with full-rank covariance and the outliers are modeled (in the latent space) by a Gaussian with a low-rank covariance. One can note that this will increase the reconstruction loss.

\subsection{Additional Implementation Details of MAW}
\label{sec:MAW_implement}

All NNs were implemented with TensorFlow (available at \url{tensorflow.org})
% \citep{tensorflow2015-whitepaper} 
and trained for $100$ epochs with batch size $128$. We apply batch normalization to each layer of any NN. For the VAE-structure of MAW, we use Adam with a learning rate of $0.0005$. For the WGAN-type discriminator of MAW, we perform RMSprop \citep{bengio2005non} with a learning rate of $0.0005$, following the recommendation of \citet{arjovsky2017wasserstein} for WGAN. 
For all experiments, the MAW discriminator is a fully connected network of size $(32, 64, 128, 1)$. The matrix $\mA$ and the network parameters for encoders, decoders and discriminators are initialized by the Glorot uniform initializer \citep{glorot2010understanding}.

The implementation details of the reconstruction-based methods are similar to those of MAW. In particular, we optimized using Adam \citep{kingma2014adam} with a learning rate of $0.0005$.

\subsection{Algorithms for MAW}
\label{sec:alg_box}
Algorithms \ref{alg:the_alg} and \ref{alg:the_alg_test}
describe the training and application of MAW for novelty detection.
We denote by $\vtheta$, $\vvarphi$ and $\vvardelta$ the trainable parameters of the encoder $\gE$, decoder $\gD$ and discriminator $\Dis$, respectively. Recall that $\mA$ includes the trained parameters of the dimension reduction component.

\begin{algorithm}[h!] 
 \caption{Training MAW}
 \label{alg:the_alg}
 \begin{algorithmic}[1]
 \renewcommand{\algorithmicrequire}{\textbf{Input:}}
 \renewcommand{\algorithmicensure}{\textbf{Output:}}
 \REQUIRE Training data $\{\rvx^{(i)}\}_{i=1}^L$; initialized parameters $\vtheta$, $\vvarphi$ and $\vvardelta$ of $\gE$, $\gD$ and $\Dis$, respectively; initialized $\mA$; weight $\eta$; number of epochs; batch size $I$; sampling number $T$; learning rate $\alpha$
 \ENSURE Trained parameters $\vtheta$, $\vvarphi$ and $\mA$
  \FOR {each epoch}
  \FOR {each batch $\{\rvx^{(i)}\}_{i \in \gI}$}
  \STATE $\vmu_{0, 1}^{(i)}, \vmu_{0, 2}^{(i)}, \vs_{0, 1}^{(i)}, \vs_{0, 2}^{(i)} \gets \gE( \rvx^{(i)} )$
  \STATE $\vmu_j^{(i)} \gets \mA^\T \vmu_{0, j}^{(i)},~\mM_j^{(i)} \gets \mA^\T \textup{diag}(\vs_{0, j}^{(i)}) \mA, ~j = 1,2$
  \STATE Compute $\tilde{\mM}_1^{(i)}$ according to \eqref{eq:decompose_M1} and \eqref{eq:synthesis_M1}
  \STATE $\mSigma_1^{(i)} \gets \tilde{\mM}_1^{(i)} \tilde{\mM}_1^{(i)\T},~\mSigma_2^{(i)} \gets \mM_2^{(i)} \mM_2^{(i)\T}$
    \FOR {$t = 1, \cdots, T$}
    \STATE sample a batch $\{\rvz_{\textup{gen}}^{(i,t)}\}_{i \in \gI} \sim \eta \gN(\vmu_1^{(i)}, \mSigma_1^{(i)}) + (1-\eta) \gN(\vmu_2^{(i)}, \mSigma_2^{(i)})$
    \STATE sample a batch $\{\rvz_{\textup{hyp}}^{(i,t)}\}_{i \in \gI} \sim \gN(\bm{0},\mI)$
    \ENDFOR
  \STATE $(\vtheta, \mA, \vvarphi) \gets (\vtheta, \mA, \vvarphi) - \alpha \nabla_{(\vtheta, \mA, \vvarphi)} L_{\rm{VAE}}(\vtheta, \mA, \vvarphi)$ according to \eqref{eq:vae_loss}
  \STATE $\vvardelta \gets \vvardelta - \alpha \nabla_\vvardelta L_{W_{1}}(\vvardelta)$ according to \eqref{eq:wass_loss}
  \STATE $\vvardelta \gets \textup{clip}(\vvardelta, [-1, 1])$
  \STATE $(\vtheta, \mA) \gets (\vtheta, \mA) - \alpha \nabla_{(\vtheta, \mA)} L_{\rm{GEN}}(\vtheta, \mA)$ according to \eqref{eq:enc_loss}
  \ENDFOR
  \ENDFOR
 \end{algorithmic} 
 \end{algorithm}

 \begin{algorithm}[ht!] 
 \caption{Applying MAW to novelty detection}
 \label{alg:the_alg_test}
 \begin{algorithmic}[1]
 \renewcommand{\algorithmicrequire}{\textbf{Input:}}
 \renewcommand{\algorithmicensure}{\textbf{Output:}}
 \REQUIRE Test data $\{\rvy^{(j)}\}_{j=1}^{N}$; sampling number $T$; trained MAW model; threshold $\rvepsilon_{\rm{T}}$;  similarity $S(\cdot,\cdot)$
 \ENSURE Binary labels for novelty for each $j = 1, \ldots, N$
  \FOR{$j = 1, \ldots, N$}
  \STATE $\vmu_{0, 1}^{(j)}, \vs_{0, 1}^{(j)} \gets \gE( \rvy^{(j)} )$
  \STATE $\vmu_1^{(j)} \gets \mA^\T \vmu_{0, 1}^{(j)},~\mM_1^{(j)} \gets \mA^\T \textup{diag}(\vs_{0, 1}^{(j)}) \mA$
  \STATE Compute $\tilde{\mM}_1^{(j)}$ according to \eqref{eq:decompose_M1} and \eqref{eq:synthesis_M1}
  \STATE $\mSigma_1^{(j)} \gets \tilde{\mM}_1^{(j)} \tilde{\mM}_1^{(j)\T}$
  \FOR{$t = 1, \cdots, T$}
  \STATE sample $\rvz_{\textup{in}}^{(j,t)} \sim \gN(\vmu_1^{(j)}, \mSigma_1^{(j)})$
  \STATE $\tilde{\rvy}^{(j, t)} \gets \gD \left( \rvz_{\textup{in}}^{(j,t)} \right)$
  \STATE compute $S(\rvy^{(j)}, \tilde{\rvy}^{(j, t)})$
  \ENDFOR
  \STATE $S^{(j)} \gets T^{-1} \sum_{t=1}^T S(\rvy^{(j)}, \tilde{\rvy}^{(j,t)})$
  \IF {$S^{(j)} \geq \rvepsilon_{\rm{T}}$}
  \STATE $\rvy^{(j)}$ is a normal example
  \ELSE
  \STATE $\rvy^{(j)}$ is a novelty
  \ENDIF
 \ENDFOR
 \end{algorithmic} 
 \end{algorithm}

\section{SENSITIVITY TO SOME HYPERPARAMETERS}\label{sec:sensitivity}
We examine sensitivity to choices of the intrinsic dimension (see \S\ref{sec:diffdim}) and the mixture parameter (see \S\ref{sec:diffmix}).
\subsection{Sensitivity to the Intrinsic Dimension}\label{sec:diffdim} 
\begin{figure}[ht!]
\centering
\includegraphics[width=0.75\columnwidth]{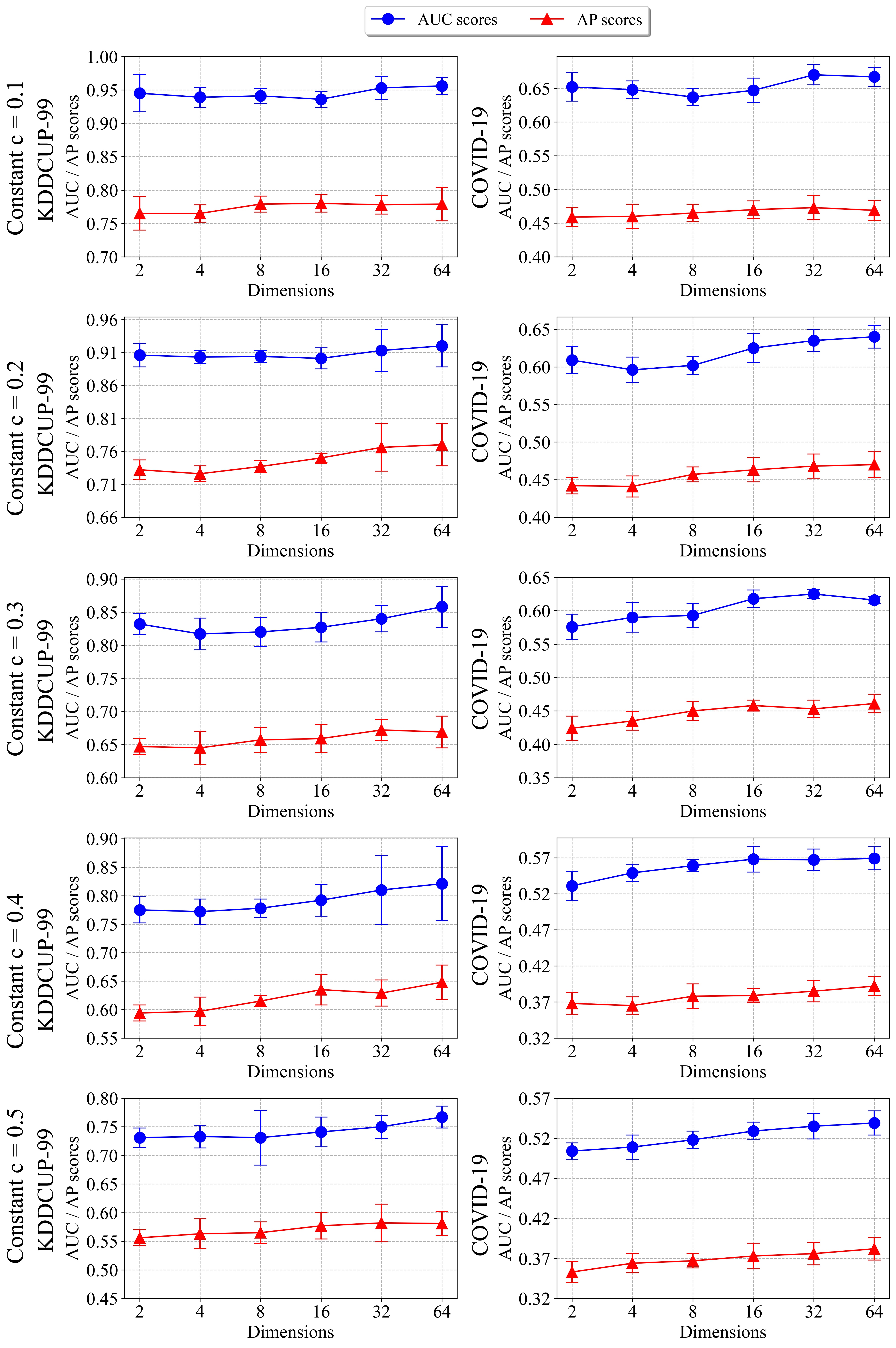}
\caption{AUC and AP scores with intrinsic dimensions $d=2$, $4$, $8$, $16$, $32$ and $64$ for KDDCUP-99 (on the left) and COVID-19 (on the right), where $c \in \{0.1, 0.2, 0.3, 0.4, 0.5 \}$}. 
\label{fig:diffdims}
\end{figure}

Our default value of the intrinsic dimension is $d=2$. 
Here we study the sensitivity of our numerical results to the following choices intrinsic dimensions: $d=2$, $4$, $8$, $16$, $32$ and $64$, while using the KDDCUP-99 and COVID-19 datasets.  
The training ratio of outliers per inliers $c$ are in $\{0.1, 0.2, 0.3, 0.4, 0.5 \}$.
We compute the AUC and AP scores averaged over the testing ratios of outliers per inliers, $c_{\textup{test}}=0.1$, $0.3$, $0.5$, $0.7$ and $0.9$, and over three runs of the same setting. Fig.~\ref{fig:diffdims} reports the averaged results and their standard deviations, which are indicated by error bars.

We can see that larger intrinsic dimensions generally result in better performances. However, the improvement is not significant and not consistent for smaller dimensions. Furthermore, higher dimensions require more substantial computational efforts for training. 

\subsection{Sensitivity to the Mixture Parameter}\label{sec:diffmix} 
\begin{figure}[th!]
\centering
\includegraphics[width=0.75\columnwidth]{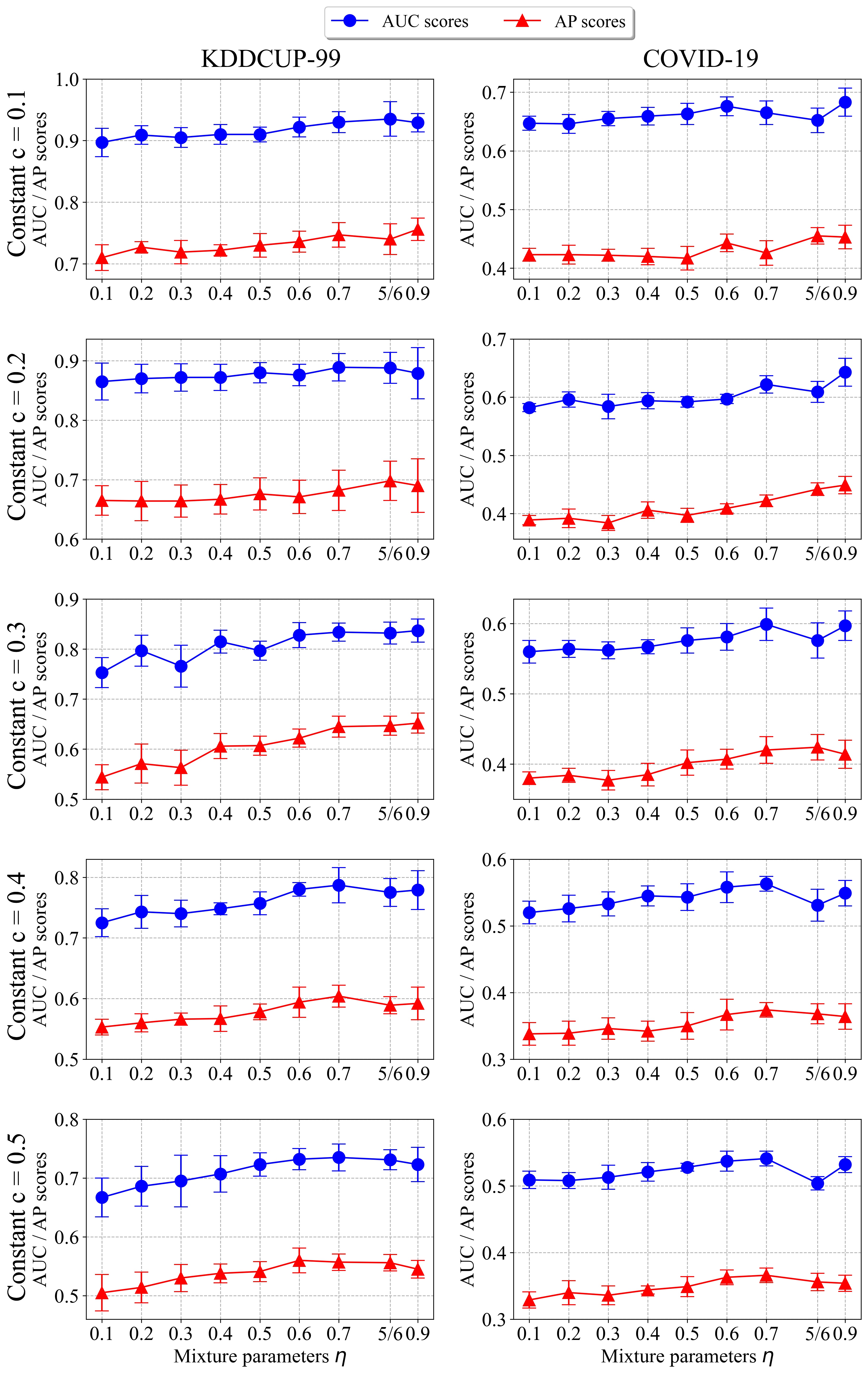}
\caption{AUC and AP scores with mixture parameters $\eta =$ $0.1$, $0.2$, $0.3$, $0.4$, $0.5$, $0.6$, $0.7$, $5/6$ and $0.9$ for KDDCUP-99 (on the left) and COVID-19 (on the right). From the top to the bottom row, the training ratios of outliers per inliers are $c=0.1$, $0.2$, $0.3$, $0.4$ and $0.5$, respectively.} 
\label{fig:diffmix}
\end{figure}

The default value of the mixture parameter $\eta$ is $5/6$. Here we study the sensitivity of the accuracy of MAW to the mixture parameters: $\{0.1, 0.2, 0.3, 0.4, 0.5, 0.6, 0.7, 5/6, 0.9 \}$.
We use $5/6 \approx 0.83$, instead of the nearby value 0.8, since it was already tested for MAW.
The training ratios of outliers per inliers are $0.1$, $0.2$, $0.3$, $0.4$ and $0.5$. 
Following the same procedure of \S\ref{sec:sensitivity}, we average the AUC and AP scores for both KDDCUP-99 and COVID-19. We report them in Fig.~\ref{fig:diffmix}.

We notice that the AUC and AP scores mildly increase as $\eta$ increases (though they may slightly decrease at 0.9). It seems that MAW learns well the inlier mode with a sufficiently large inlier weight, 
where the variation in the accuracy as a function of $\eta$ is not large in general.

\newpage
\null
\newpage
\null
\newpage

\section{RUNTIME COMPARISON}
\label{sec:runtime}
Table~\ref{tab:runtime} summarizes runtimes of all the above experiments with $c=0.3$. The initially computed runtimes are times measured for completing single experiments with a single epoch. The table averages each such runtime over the different classes and different outlier ratios for testing.
We note that LOF, OCSVM and IF are faster than the rest of the methods since they do not require training neural networks. Among the neural-networks-based methods, RSRAE is the fastest and OCGAN, DAGMM and MAW are the slowest. Indeed, RSRAE has a single autoencoder and OCGAN, DAGMM and MAW contain several neural networks. Another possible reason for the relative slowness of MAW is due to its dimension reduction component, whose implementation in TensorFlow seems to be computationally expensive.
However, it seems to help achieve competitive performance in detecting outliers. We plan to investigate a more efficient implementation of the dimension reduction component in the future.

\begin{table*}[h!]
  \caption{Runtimes (in seconds) of competing methods when the training ratio of outliers per inliers is $c=0.3$. }
  \label{tab:runtime}
  \centering
  \resizebox{0.9\textwidth}{!}{
  \begin{tabular}{lcccccc}
    \cmidrule(r){1-7}
    Methods & COVID-19 & CIFAR-10 & Caltech101  & Fashion MNIST & KDDCUP-99 & Reuters-21578\\
    \midrule
    LOF     & 0.30 $\pm$ 0.17 &  3.98 $\pm$ 0.13 &  0.24 $\pm$ 0.01 &   16.31 $\pm$ 1.01 & 3.23  $\pm$ 0.04 & 17.91 $\pm$ 1.98\\
    OCSVM   & 0.17 $\pm$ 0.07 &  2.22 $\pm$ 0.09 &  0.12 $\pm$ 0.00 & 8.34  $\pm$ 2.36 &  9.08 $\pm$ 0.05 & 8.74 $\pm$ 1.47\\
    IF     & 0.43 $\pm$ 0.01 &  1.86 $\pm$ 0.12 & 0.39  $\pm$ 0.01 & 2.86  $\pm$ 0.37 &  1.67 $\pm$ 0.03 & 10.54 $\pm$ 1.89\\
    RSRAE   & 4.31 $\pm$ 0.45 &  8.49 $\pm$ 0.77 & 5.69  $\pm$ 0.36 & 23.69  $\pm$ 0.39 & 40.18  $\pm$ 0.33 & 6.22 $\pm$ 0.25\\
    DSEBMs  & 48.30 $\pm$ 3.45 & 66.57  $\pm$ 2.35 &  147.15 $\pm$ 0.32 &  151.02 $\pm$ 7.67 & 216.02  $\pm$ 4.34 & 74.28 $\pm$ 2.09\\
    OCGAN   & 182.79 $\pm$ 2.53 &  313.28 $\pm$ 0.13 &  679.44 $\pm$ 4.62 &  250.51 $\pm$ 0.24 &  2035.83 $\pm$ 8.34 & 343.02 $\pm$  7.42\\
    DAGMM   & 99.44 $\pm$ 5.76 & 134.36  $\pm$ 9.12 &  504.11 $\pm$ 11.31 &  353.65 $\pm$ 14.57 &  396.44 $\pm$ 7.65 & 177.30 $\pm$ 3.56\\
    MAW    & 136.42 $\pm$ 0.16 &   1871.22$\pm$ 16.03 &   217.32 $\pm$ 0.25 &  1441.97 $\pm$ 15.123 &   3166.62 $\pm$ 12.12 & 255.85 $\pm$ 2.97\\    
    \bottomrule
  \end{tabular}
  } 
\end{table*}

\section{EXPERIMENTS WITH DIFFERENT TYPES OF OUTLIERS} \label{sec:mix_data}
\begin{figure}[h!]
\centering
\includegraphics[width=0.67\columnwidth]{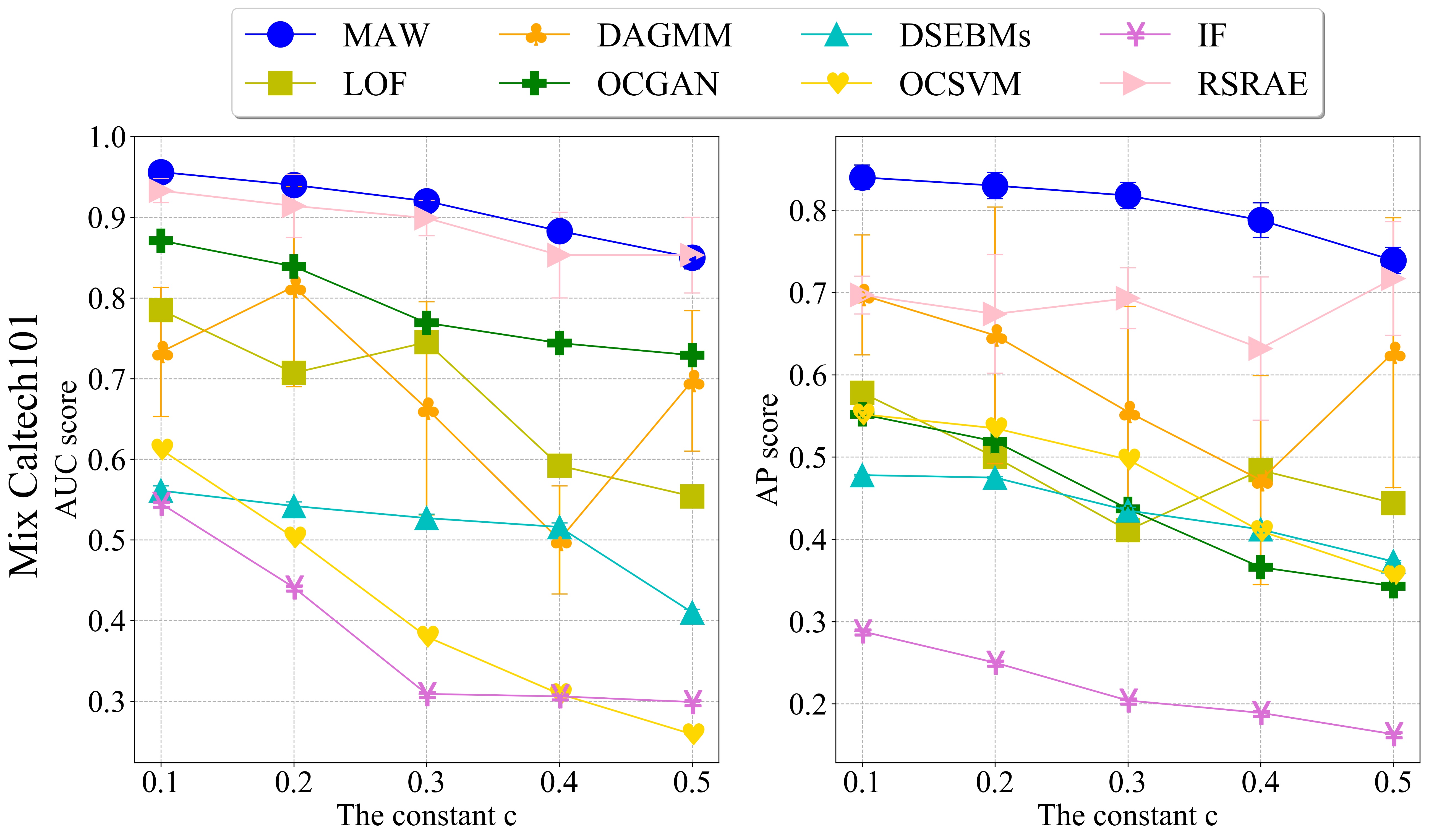}
\caption{AUC and AP scores with training ratio of outliers per inliers $c \in \{0.1, 0.2, 0.3, 0.4, 0.5 \}$ for the Mix Caltech101 dataset.} 
\label{fig:caltechmix}
\end{figure}

In this section, we test the performance of MAW and the benchmark methods when the training and test sets are corrupted by outliers with different structures. 
We generate a dataset, which we call ``Mix Caltech101'', in the following way. We fix the largest class of Caltech101 (containing airplane images) 
as the inlier class and randomly split it into the training inlier class (68.75$\%$) and testing inlier class (31.25$\%$). We form the training set by corrupting the training inlier class with random samples from the ten classes of CIFAR-10 \citep{krizhevsky2009learning} with training ratio of outliers per inliers $c \in \{0.1, 0.2, 0.3, 0.4, 0.5 \}$. For the test set, we corrupt the testing inlier class by ``tile images’’ from MVTech dataset \citep{bergmann2019mvtec} with testing ratio of outliers per inliers $c_{\textup{test}}$ 
in $\{ 0.1, 0.3, 0.5, 0.7, 0.9 \}$. The rest of the settings of the experiments are identical to the description in \S\ref{subsec:res} of the main text. We present the AUC and AP scores and their standard deviations in Fig.~\ref{fig:caltechmix}.

Clearly, the competitive advantage of MAW is also noticeable in this setting. We note that OCSVM, the traditional distance-based method, and IF, the traditional density-based method, perform poorly in this scenario, whereas they performed well in our original setting.

\section{ADDITIONAL THEORETICAL GUARANTEES FOR THE $W_1$ MINIMIZATION}\label{sec:moretheory} 

In \S\ref{subsec:motivation} we fully motivate our focus on studying \eqref{eq:bary} in order to understand the advantage of the use of the Wasserstein distance over the KL divergence in the framework of MAW. In \S\ref{subsec:proof} we prove Proposition \ref{prop:baryW1andKL}. In \S\ref{subsec:W2theory}, we discuss a possible deviation of the clean theory of Proposition~\ref{prop:baryW2} from practice. In \S\ref{proof:baryW2} we prove Proposition~\ref{prop:baryW2} and in \S\ref{subsec:proofbaryKL} we prove Proposition~\ref{prop:baryKL}.

\subsection{Motivation for Studying \eqref{eq:bary}}\label{subsec:motivation}
The implementation of any VAE or its variants, such as AAE, WAE and MAW, requires the optimization of a regularization penalty ${\cal R}$, which measures the discrepancy between the latent and prior distributions. This penalty is typically the KL divergence, though one may use appropriate metrics such as $W_2$ or $W_1$. Thus one needs to minimize 
\begin{equation}\label{eq:minagglatent}
    {\cal R} \left( \frac{1}{L} \sum_{i=1}^L q(\rvz|\rvx^{(i)}), p(\rvz) \right)
\end{equation}
over the %observed 
variational family $\gQ = \{q(\rvz|\rvx)\}$ indexed by some parameters. Here $L$ is the batch size of the input data and $\sum_{i=1}^L q(\rvz|\rvx^{(i)})$ is its observed aggregated distribution. 

Since the explicit expressions of the regularization measurements between aggregated distributions are unknown, it is not feasible to study the minimizer of \eqref{eq:minagglatent}. We thus consider the following approximation of \eqref{eq:minagglatent}: 
\begin{equation}
    \label{eq:minagglatent_2}
\sum_{i=1}^L \frac{1}{L} {\cal R} \left(q(\rvz|\rvx^{(i)}), p(\rvz) \right). \end{equation}
We can minimize one term of this sum at a time, that, is minimize ${\cal R} \left(q(\rvz|\rvx), p(\rvz) \right)$ over $\gQ$. This minimization strategy is common in the study of the Wasserstein barycenter problem \citep{agueh2010barycenters, peyre2019computational, chen2018optimal}.

One of the underlying assumptions of MAW is that the prior distribution $p(\rvz)$ is Gaussian and $q(\rvz|\rvx)$ is a Gaussian mixture. That is,  $p(\rvz) = \gN(\rvz \vert \bm{\mu_0}, \mSigma_0)$ and $q(\rvz|\rvx) = \eta \gN(\rvz \vert \vmu_1, \mSigma_1) + (1-\eta) \gN(\rvz \vert \vmu_2, \mSigma_2)$. This gives rise to the following minimization problem
\begin{align}\label{eq:minmixlatent}
    \min_{\substack{\vmu_1, \vmu_2 \in \R^K; \mSigma_1, \mSigma_2 \in {\cal S}_{+}^{K}}}
 & {\cal R}  \biggl( \eta \gN(\vmu_1, \mSigma_1) 
 + (1-\eta) \gN(\vmu_2, \mSigma_2),  \gN(\vmu_0, \mSigma_0) \biggr).
\end{align} 
Similarly to approximating  \eqref{eq:minagglatent}
by \eqref{eq:minagglatent_2}, we approximate \eqref{eq:minmixlatent} by \eqref{eq:bary}. We remark that in \eqref{eq:bary} we further assume that there is a sufficiently small threshold $\epsilon > 0$ for which $\norm{\vmu_1 - \vmu_2}_2 \geq \epsilon$. This is a reasonable assumption since, in practice, if $\vmu_1$ and $\vmu_2$ are very close, the reconstruction loss will be large.

\subsection{Proof of Proposition~\ref{prop:baryW1andKL}}\label{subsec:proof}

Recall that $\vmu_0  \in \R^K$ is the mean of the prior Gaussian, $\epsilon >0$ is the fixed separation parameter for the means of the two modes and $\eta>1/2$ is the fixed mixture parameter. 
For $i=0$, 1, 2, we denote the Gaussian probability distribution by $\nu_i=\mathcal{N} (\vmu_i, \mSigma_i)$. Since in our setting $\mSigma_0 = \mSigma_1 = \mSigma_2$, we denote the common covariance matrix in ${\cal S}_{++}^{K}$ by $\mSigma$.
That is, $\mSigma = \mSigma_i$ for $i=0,1,2$.

We first analyze the solution of \eqref{eq:bary} with ${\cal R} = W_p$, where $p \geq 1$, and then analyze the solution of \eqref{eq:bary} with ${\cal R} = KL$.

{\bf The case} $\boldsymbol{{\cal R} = W_p,  p \geq 1}${\bf :}
We follow the next three steps to prove that the minimizer of \eqref{eq:bary} satisfies $\vmu_1=\vmu_0$.

{\bf Step I:} We prove that 
\begin{equation}\label{eq:meanW1}
 \begin{split} 
     W_p(\nu_i, \nu_0) &\equiv W_p(\gN(\vmu_i, \mSigma), \gN(\vmu_0, \mSigma))
     \\&= \norm{\vmu_i - \vmu_0}_2 \ \text{ for } p \geq 1 \text{ and } i=1,2~.
\end{split}   
\end{equation}

First, we note that using the definition of $W_p, p \geq 1$ and the common notation $\Pi(\nu_i, \nu_0)$ for the distribution on $\R^K \times \R^K$ with marginals $\nu_i$ and $\nu_0$
\begin{equation}
\label{eq:Wp_lower_bound}
\begin{split}
    W_p^p(\nu_i, \nu_0) 
    &=  \inf_{\pi \in \Pi(\nu_i, \nu_0)} \E_{(\rvx,\rvy) \sim \pi} \norm{\rvx-\rvy}_2^p \\
    & \geq \inf_{\pi \in \Pi(\nu_i, \nu_0)}  \norm{\E_{(\rvx,\rvy) \sim \pi}\rvx- \E_{(\rvx,\rvy) \sim \pi}\rvy}_2^p \\
    & = \norm{\vmu_i - \vmu_0}_2^p ~,
\end{split}
\end{equation} 
where the inequality follows from the fact that $\norm{.}_2^p$ is convex and from Jensen's inequality.

On the other hand, for $i=1$ or $i=2$,
let $\rvx^*$ be an arbitrary random vector with distribution $\nu_i$, and let $\rvy^* = \rvx^* - \vmu_i + \vmu_0$. The distribution of $\rvy^*$ is Gaussian with mean $\vmu_0$ and covariance $\mSigma_i$, that is, this distribution is $\nu_0$. Let $\pi^*$ be the joint distribution of the random variables $\rvx^*$ and $\rvy^*$. We note that $\pi^*$ is in $\Pi(\nu_i, \nu_0)$ and that 
$$
    \E_{(\rvx,\rvy) \sim \pi^*} \norm{\rvx - \rvy}_2^p  = \E_{(\rvx,\rvy) \sim \pi^*} \norm{\vmu_i - \vmu_0}_2^p = \norm{\vmu_i - \vmu_0}_2^p ~.
$$
Therefore, 
\begin{equation}
\label{eq:Wp_upper_bound}
\begin{split}
    W_p^p(\nu_i, \nu_0) &= \inf_{\pi \in \Pi(\nu_i, \nu_0)} \E_{(\rvx,\rvy) \sim \pi} \norm{\rvx-\rvy}_2^p 
    \\&\leq \E_{(\rvx,\rvy) \sim \pi^*} \norm{\rvx - \rvy}_2^p  = \norm{\vmu_i - \vmu_0}_2^p ~.    
\end{split}
\end{equation}
The combination of \eqref{eq:Wp_lower_bound} and \eqref{eq:Wp_upper_bound} immediately yields 
\eqref{eq:meanW1}.

{\bf Step II:} We prove that \eqref{eq:bary} with ${\cal R} = W_p$, {$p \geq 1$}, is equivalent to
\begin{equation}\label{eq:colinearW1}
 \begin{split}
&\min_{\substack{\vmu_1, \vmu_2 \in \R^K;
\\ \rm{s.t.}~ \vmu_0, \vmu_1,  \vmu_2: \textup{colinear} \\ 
\& \norm{\vmu_1 - \vmu_2 }_2 \geq \epsilon}} 
\quad  \eta \norm{\vmu_1 - \vmu_0}_2 + (1-\eta) \norm{\vmu_2 - \vmu_0}_2.
\end{split}   
\end{equation}
We first note that \eqref{eq:bary} with ${\cal R} = W_p$, {$p \geq 1$}
is equivalent to
\begin{equation}\label{eq:colinearW1_2}
 \begin{split}
 &\min_{\substack{\vmu_1, \vmu_2 \in \R^K \\ \rm{s.t.}~ \norm{\vmu_1 - \vmu_2 }_2 \geq \epsilon}} 
\quad  \eta \norm{\vmu_1 - \vmu_0}_2 + (1-\eta) \norm{\vmu_2 - \vmu_0}_2.
\end{split}   
\end{equation}
Indeed, this is a direct consequence of the expression derived in step I for $\cal R$ in this case. It is thus left to show that if $\vmu_1'$, $\vmu_2' \in \R^K$ minimize \eqref{eq:colinearW1_2}, then we can construct $\widetilde{\vmu_1'}$, $\widetilde{\vmu_2'} \in \R^K$ that are colinear with $\vmu_0$ and also minimize \eqref{eq:colinearW1_2}.

\begin{figure}[h]
\centering
\includegraphics[width=0.5\columnwidth]{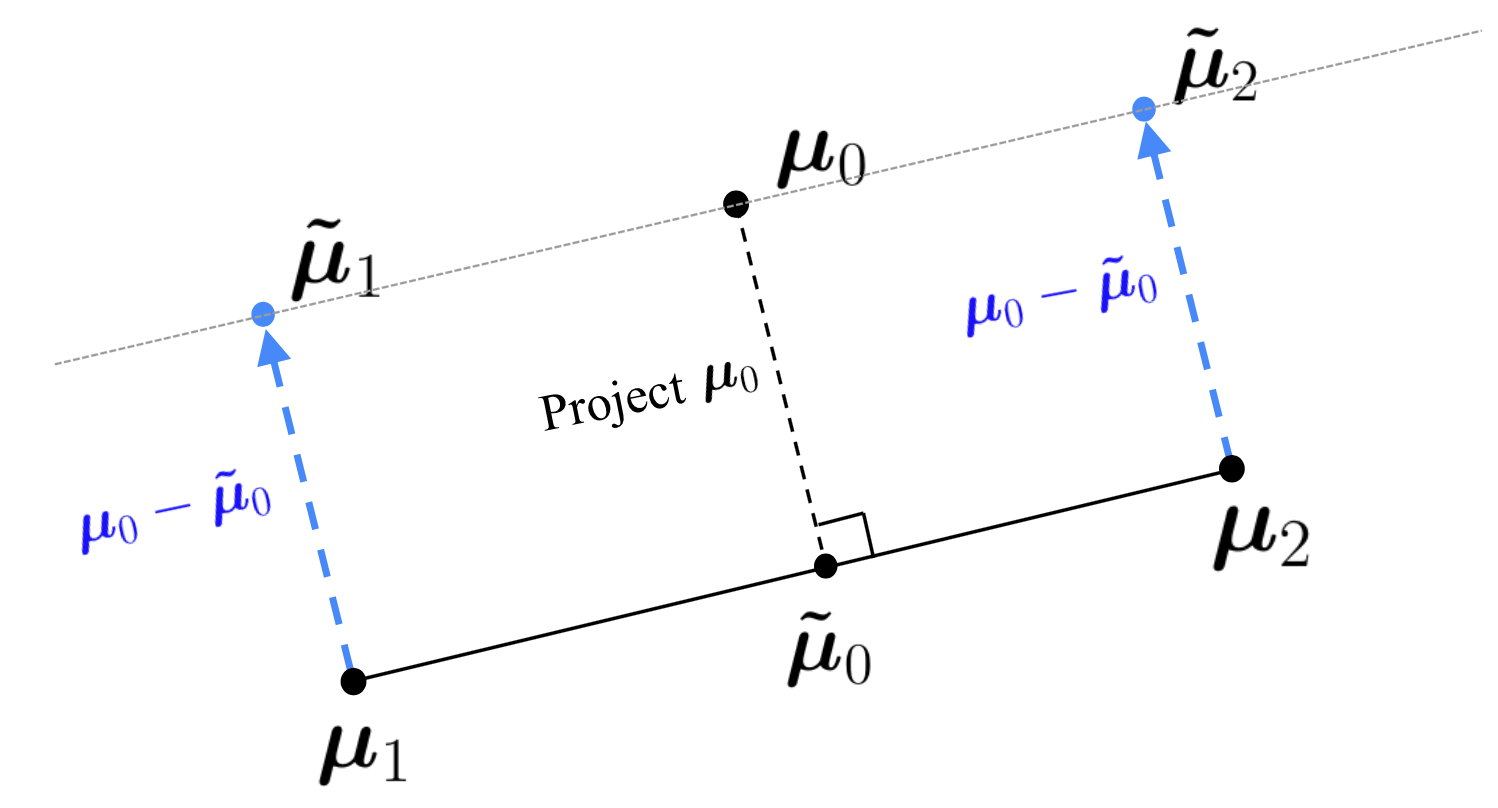} 
\caption{Illustration of the points $\bm{\tilde{\mu}}_0$, $\bm{\tilde{\mu}}_1$ and $\bm{\tilde{\mu}}_2$ and their properties.}
\label{fig:colinear}
\end{figure}

For any $\bm{\mu_1}$ and $\bm{\mu_2}$ in $\R^K$ with $\norm{\vmu_1 - \vmu_2 }_2 \geq \epsilon$ and for the given $\bm{{\mu}}_0 \in \R^K$, we define
$\bm{\tilde{\mu}}_0$, $\bm{\tilde{\mu}}_1$ and $\bm{\tilde{\mu}}_2 \in \R^K$ and demonstrate them in Fig.~\ref{fig:colinear}. The point $\bm{\tilde{\mu}}_0$ is the projection of $\vmu_0$ onto $\vmu_1 - \vmu_2$ and $\bm{\tilde{\mu}}_i := \vmu_i + \vmu_0 - \bm{\tilde{\mu}}_0$ for $i=1$, $2$.
We observe the following properties, which can be proved by direct calculation, though Fig.~\ref{fig:colinear} also clarifies them:
\begin{equation*}
\norm{\vmu_i - \vmu_0}_2 \geq \norm{\bm{\tilde{\mu}}_i - \vmu_0}_2 
\text{ for } i=1, 2,    
\end{equation*}
and consequently,
\begin{equation}
\label{eq:prop1}
% \begin{split}
      \eta \norm{\vmu_1 - \vmu_0}_2 + (1-\eta) \norm{\vmu_2 - \vmu_0}_2  
      \geq \eta \norm{\bm{\tilde{\mu}}_1 - \vmu_0}_2 + (1-\eta) \norm{\bm{\tilde{\mu}}_2 - \vmu_0}_2; 
% \end{split}
\end{equation}
\begin{equation}
\label{eq:prop2}
\norm{\bm{\tilde{\mu}}_1 - \bm{\tilde{\mu}}_2}_2 = \norm{\vmu_1-\vmu_2}_2 \geq \epsilon;    
\end{equation}
and
\begin{equation}
\label{eq:prop3}
\bm{\tilde{\mu}}_1, \ \bm{\tilde{\mu}}_2, \ \text{ and } \vmu_0 \ \text{ are colinear.}
\end{equation}
Clearly, the combination of \eqref{eq:prop1}, \eqref{eq:prop2} and \eqref{eq:prop3} concludes the proof of step II. That is, it implies that if $\vmu_1'$, $\vmu_2' \in \R^K$ minimize \eqref{eq:colinearW1_2}, then $\widetilde{\vmu_1'}$ and $\widetilde{\vmu_2'}$ defined above are colinear with $\vmu_0$ and also minimize \eqref{eq:colinearW1_2}.

{\bf Step III:} We directly solve \eqref{eq:colinearW1} and consequently 
\eqref{eq:bary} with ${\cal R} = W_p$, $p \geq 1$. Due to the  colinearity constraint in \eqref{eq:bary}, we can write 
\begin{equation}
\label{eq:colinear_t}
\bm{\mu_0} = (1+t) \bm{\mu_1} - t\bm{\mu_2} \ \text{ for } t \in \R.    
\end{equation}
The objective function in \eqref{eq:colinearW1} can then be written as 
\begin{equation*}
% \begin{split}
        \norm{\vmu_1 - \vmu_2 }_2 \left ( \eta |t| + (1-\eta) |1+t| \right ) 
        \geq \epsilon \left ( \eta |t| + (1-\eta) |1+t| \right ),
% \end{split}
\end{equation*}
where equality is achieved if and only if $\norm{\vmu_1-\vmu_2}_2 =\epsilon$.
We thus define $r(t) = \eta |t| + (1-\eta) |1+t| $ and note that  
\begin{equation*}\label{eq:derW1}
r(t) = \begin{cases} 
t + (1-\eta), & t\geq 0 \\ 
(1-2\eta)t + (1-\eta), & 0\geq t\geq -1 \\ 
-t + (\eta -1), & -1\geq t \end{cases}
\end{equation*}
and its derivative is
\begin{equation*}
r'(t) = \begin{cases} 
1, & t>0 \\ 
1-2\eta, & 0>t>-1 \\ 
-1, & -1>t. \end{cases}
\end{equation*}
The above expressions for $r$ and $r'$ and the assumption that $\eta>1/2$ imply that $r(t)$ is increasing when $t>0$, decreasing when $t<0$ and $r(0) = 1- \eta < \eta = r(1)$. Thus $r$ has a global minimum at $t=0$. Hence, it follows from \eqref{eq:colinear_t} that the minimizer of \eqref{eq:bary}, and equivalently \eqref{eq:bary} with ${\cal R} = W_p$, $p \geq 1$ satisfies $\vmu_1 = \vmu_0$.

{\bf The case} $\boldsymbol{{\cal R} = KL}${\bf :}
We  prove that the solution of \eqref{eq:bary} with ${\cal R} = KL$ satisfies $\vmu_0 = \eta \vmu_1 + (1-\eta) \vmu_2$. We practically follow similar steps as the proof above.

{\bf Step I:} We derive an expression for $KL(\nu_i || \nu_0)$, where $i=1$, 2.
We use the following general formula, which holds for the case where $\mSigma_0$, $\mSigma_1$ and $\mSigma_2$ are general covariance matrices in ${\cal S}_{++}^{K}$ (see e.g., (2) in \citep{hershey2007approximating}): 

\begin{equation}\label{eq:closedKL}
% \begin{split}
        KL(\nu_i || \nu_0) = \dfrac{1}{2} \biggl( \log \dfrac{\det \mSigma_0 }{\det \mSigma_i} -K + \textup{tr}(\mSigma_0^{-1}\mSigma_i) 
        + (\vmu_i - \vmu_0)^{\textup{T}} \mSigma_0^{-1} (\vmu_i - \vmu_0) \biggr)~.
% \end{split}
\end{equation}

Since in our setting $\mSigma_1 = \mSigma_2 = \mSigma$, 
this expression has the simpler form:
\begin{equation*}\label{eq:samecovKL}
    KL(\nu_i || \nu_0) = \dfrac{1}{2}  (\vmu_i - \vmu_0)^{\textup{T}} \mSigma^{-1} (\vmu_i - \vmu_0).
\end{equation*}

{\bf Step II:} We reformulate the optimization problem.
The above step implies that \eqref{eq:bary} with ${\cal R} = KL$ can be written as
\begin{equation*}\label{eq:equ1KL_0}
%  \begin{split}
 \min_{\substack{ \norm{\vmu_1 - \vmu_2 }_2 \geq \epsilon}} 
\quad  \eta (\vmu_1 - \vmu_0)^{\textup{T}} \mSigma^{-1} (\vmu_1 - \vmu_0)
 + (1-\eta) (\vmu_2 - \vmu_0)^{\textup{T}} \mSigma^{-1} (\vmu_2 - \vmu_0),
% \end{split}   
\end{equation*}
or equivalently,
\begin{equation}\label{eq:equ1KL}
%  \begin{split}
 \min_{\substack{ \norm{\vmu_1 - \vmu_2 }_2 \geq \epsilon}} 
\quad  \eta \norm{ \mSigma^{-\frac{1}{2}} (\vmu_1 - \vmu_0)}_2^2 + 
(1-\eta) \norm{ \mSigma^{-\frac{1}{2}} (\vmu_2 - \vmu_0)}_2^2.
% \end{split}   
\end{equation}
We express the eigenvalue decomposition of $\mSigma^{-1}$ as
$\mSigma^{-1} = \bm{U} \bm{\Lambda} \bm{U}^{\textup{T}}$, where $\bm{\Lambda} \in {\cal S}_{+}^{K}$, 
and $\bm{U}$ is an orthogonal matrix. Applying the change of variables 
$\vmu_i^{'} = \bm{\Lambda}^{\frac{1}{2}}  \bm{U}^{\textup{T}} \vmu_i$
for $i = 0$, 1, 2,    
we rewrite \eqref{eq:equ1KL} as
\begin{equation}\label{eq:equ2KL_0}
 \begin{split}
 &\min_{\substack{  \norm{ %\bm{\Lambda}^{-\frac{1}{2}}  
 \vmu_1^{'} - %\bm{\Lambda}^{-\frac{1}{2}}  
 \vmu_2^{'}  }_2 \geq \epsilon}}
\quad  \eta \norm{ %\bm{\Lambda}^{-\frac{1}{2}}   
\vmu_1^{'} -%\bm{\Lambda}^{-\frac{1}{2}}  
\vmu_0^{'}}_2^2 + (1-\eta) \norm{ %\bm{\Lambda}^{-\frac{1}{2}} 
\vmu_2^{'} - %\bm{\Lambda}^{-\frac{1}{2}}  
\vmu_0^{'}}_2^2.
%\\ = & \min_{\substack{ \substack{ \vmu_0^{'}, \vmu_1^{'}, \vmu_2^{'}: ~\textup{colinear}
%\\ \& \norm{\bm{\Lambda}^{-\frac{1}{2}} \vmu_1^{'} - \bm{\Lambda}^{-\frac{1}{2}} \vmu_2^{'} }_2 =\epsilon}}}
%\quad  \eta \norm{\bm{\Lambda}^{-\frac{1}{2}} \vmu_1^{'} - \bm{\Lambda}^{-\frac{1}{2}} \vmu_0^{'}}_2^2 + (1-\eta) \norm{\bm{\Lambda}^{-\frac{1}{2}} \vmu_2^{'} - \bm{\Lambda}^{-\frac{1}{2}} \vmu_0^{'}}_2^2
\end{split}   
\end{equation}
At last, applying the same colinearity argument as above (supported by Fig.~\ref{fig:colinear}) 
%and the relationship
%$\bm{\Lambda}^{-\frac{1}{2}} \vmu_i^{'} = \bm{U}^{\textup{T}} \vmu_i$, $i = 0$, 1, 2, which follows from \eqref{eq:change_var}, 
we conclude the following equivalent formulation of \eqref{eq:equ2KL_0}:
\begin{equation}\label{eq:equ2KL}
 \begin{split}
% &\min_{\substack{  \norm{ \bm{\Lambda}^{-\frac{1}{2}}  \vmu_1^{'} - \bm{\Lambda}^{-\frac{1}{2}}  \vmu_2^{'}  }_2 =\epsilon}}
%\quad  \eta \norm{ \bm{\Lambda}^{-\frac{1}{2}}   \vmu_1^{'} -\bm{\Lambda}^{-\frac{1}{2}}  \vmu_0^{'}}_2^2 + (1-\eta) \norm{ \bm{\Lambda}^{-\frac{1}{2}} \vmu_2^{'} - \bm{\Lambda}^{-\frac{1}{2}}  \vmu_0^{'}}_2^2.
%\\ = 
& \min_{\substack{ \substack{ \vmu_0^{'}, \vmu_1^{'}, \vmu_2^{'} ~\textup{are colinear}
\\ \& \norm{ \vmu_1^{'} -  \vmu_2^{'} }_2 \geq \epsilon}}}
\quad  \eta \norm{ \vmu_1^{'} -  \vmu_0^{'}}_2^2 + (1-\eta) \norm{\vmu_2^{'} - \vmu_0^{'}}_2^2
\end{split}   
\end{equation}

{\bf Step III:} We directly solve \eqref{eq:equ2KL}. Due to the colinearity constraint, we can write 
\begin{equation}
\label{eq:colinear_t_2}
\bm{\mu_0}^{'} = (1+t) \bm{\mu_1}^{'} - t\bm{\mu_2}^{'} \ \text{ for } t \in \R    
\end{equation}
and express the objective  function of \eqref{eq:equ2KL} as  
\begin{align*}
  & \norm{\vmu_1^{'} - \vmu_2^{'} }_2^2 \left ( \eta t^2 + (1-\eta) (1+t)^2 \right) 
   \geq \epsilon^2 \left ( \eta t^2 + (1-\eta) (1+t)^2 \right),
\end{align*}
    
where equality is achieved if and only if $\norm{\vmu'_1-\vmu'_2}_2 =\epsilon$.
We thus define $r(t) =  \eta t^2 + (1-\eta) (1+t)^2 $ and note that $r'(t) = 2(t+ (1-\eta))$ and $r''(t) = 2$, and thus conclude that $r(t)$ obtains its global minimum at $t = \eta - 1$. This observation and \eqref{eq:colinear_t_2} imply that the minimizers $\bm{\mu_1}$ and $\bm{\mu_2}$ of \eqref{eq:bary} with ${\cal R} = KL$ satisfy $\bm{\mu_0} = \eta \bm{\mu_1} + (1-\eta) \bm{\mu_2}$.

\subsection{Some Remarks on Proposition~\ref{prop:baryW2}}
\label{subsec:W2theory}
We clarify why the statement and proof of the proposition are not sufficient for explaining the effect of the $W_1$ optimization on MAW. 
We note that the inlier and outlier covariances, $\mSigma_1$ and $\mSigma_2$, obtained by Proposition~\ref{prop:baryW2}, are diagonal. Furthermore, the proof of Proposition~\ref{prop:baryW2} clarifies that the underlying minimization problem of this proposition may assume without loss of generality that  the inlier and outlier covariances are diagonal (see e.g., \eqref{eq:mindiagW2}). On the other hand, the  numerical results in \S\ref{subsec:var} of the main text support the use of full covariances, instead of diagonal covariance. Nonetheless, we claim that the full covariances of MAW come naturally from the dimension reduction component of MAW. This component also contains trainable parameters for the covariances and they will affect the weights of the encoder, that is, will affect both the $W_1$ minimization and the reconstruction loss. Thus the analysis of the $W_1$ minimization component is not sufficient for inferring the whole behavior of MAW. For tractability purposes, the minimization in 
\eqref{eq:bary} ignores the dimension reduction component. For completeness we remark that there are two other differences between the use of \eqref{eq:bary} in Proposition~\ref{prop:baryW2} and the way it arises in MAW that may possibly also result in the advantage of using full covariance in MAW. First of all, the minimization in Proposition~\ref{prop:baryW2} uses ${\cal R} = W_2$, whereas MAW uses ${\cal R} = W_1$, which we find intractable when using the rest of the setting of Proposition~\ref{prop:baryW2}. Second of all, \eqref{eq:bary} with ${\cal R} = W_1$ is an approximation of the minimization of $W_1\left( \frac{1}{L} \sum_{i=1}^L q(\rvz|\rvx^{(i)}), p(\rvz)  \right)$ (see \S\ref{subsec:motivation} for explanation), which is also intractable (even if one uses ${\cal R} = W_2$).

\subsection{Proof of Proposition~\ref{prop:baryW2}}
\label{proof:baryW2}
We follow the same steps of the proof of Proposition~\ref{prop:baryW1andKL}.

{\bf Step I:} We immediately verify the formula
\begin{equation}\label{eq:commuteW2}
% \begin{split}
    W_2(\mathcal{N} (\vmu_i, \mSigma_i), \mathcal{N} (\bm{0}, \mI)) 
    = \sqrt{\norm{\vmu_i}_2^2 + \norm{\mSigma_i ^ {\frac{1}{2}} - \mI}_F^2} \ \text{ for } i = 1, 2.    
% \end{split}
\end{equation}
We use the following general formula, which holds for the case where $\mSigma_0$, $\mSigma_1$ and $\mSigma_2$ are general covariance matrices in ${\cal S}_{+}^{K}$ (see e.g., (4) in \citep{panaretos2019statistical}): For $i=1$, $2$
 \begin{equation*}
\label{eq:W2}
         W_2^2(\mathcal{N} (\vmu_i, \mSigma_i), \mathcal{N} (\vmu_0, \mSigma_0)) 
          = \norm{\vmu_i - \vmu_0}_2^2 + \textup{tr}(\mSigma_i + \mSigma_0  -2 (\mSigma_i ^ {\frac{1}{2}} \mSigma_0  \mSigma_i ^ {\frac{1}{2}} ) ^{\frac{1}{2}})~. 
 \end{equation*}
Indeed, \eqref{eq:commuteW2} is obtained as a direct consequence of \eqref{eq:W2} using the identity 
$$\textup{tr}\left(\mSigma_i + \bm{I} - 2\mSigma_i ^{\frac{1}{2}} \right) = \textup{tr}\left( \left(\mSigma_i ^{\frac{1}{2}} - \bm{I}\right)^2 \right) = \norm{\mSigma_i ^ {\frac{1}{2}} - \mI}_F^2. $$

{\bf Step II:}
We reformulate the underlying minimization problem in two different stages.
We first claim that  the minimizer of \eqref{eq:bary} with ${\cal R} = W_2$ and the constraint that $\mSigma_1$ is of rank $\kappa$ and $\mSigma_2$ is of rank $K$ can be expressed as the minimizer of 
\begin{equation}\label{eq:mindiagW2}
% \begin{split}
\min_{\substack{\vmu_1, \vmu_2 \in \R^K \rm{s.t.}~\norm{\vmu_1 - \vmu_2 }_2 =\epsilon,
\\ \mSigma_1, \mSigma_2 ~\textup{ diagonal in } \R^{K \times K}
\\ \& ~\textup{rank}(\mSigma_1) = \kappa,~\textup{rank}(\mSigma_2) = K}}
\quad \Biggl[ \eta \sqrt{\norm{\vmu_1}_2^2 + \norm{\mSigma_1 ^ {\frac{1}{2}} - \mI}_F^2} 
+ (1-\eta) \sqrt{\norm{\vmu_2}_2^2 + \norm{\mSigma_2 ^ {\frac{1}{2}} - \mI}_F^2}~ \Biggr].    
% \end{split}
\end{equation}

In view of \eqref{eq:bary} and \eqref{eq:commuteW2} we only need to prove  that the minimizer of \eqref{eq:mindiagW2} is the same if one removes the constraint that $\mSigma_1$ and $\mSigma_2$ are both diagonal matrices and require instead that they are in $\in {\cal S}_{+}^{K}$.
This is easy to show. Indeed, if for $i=1$ or $i=2$, $\mSigma_i \in {\cal S}_{+}^{K}$, then it can be diagonalized as follows: $\mSigma_i = \bm{U}_i^{\T} \bm{\Lambda}_i \bm{U}_i$, where $\bm{\Lambda}_i \in {\cal S}_{+}^{K}$ is diagonal and $\bm{U}_i$ is orthogonal. Hence, $\mSigma_i^{\frac{1}{2}} = \bm{U}_i^{\T} \bm{\Lambda}_i^{\frac{1}{2}} \bm{U}_i$ and $$\norm{\mSigma_i ^ {\frac{1}{2}} - \mI }_F^2 = \norm{\bm{U}_i^{\T} \bm{\Lambda}_i^{\frac{1}{2}} \bm{U}_i  - \mI }_F^2 = \norm{\bm{U}_i^{\T} (\bm{\Lambda}_i^{\frac{1}{2}}   - \mI) \bm{U}_i}_F^2 = \norm{\bm{\Lambda}_i^{\frac{1}{2}}-\mI}_F^2.$$ Consequently,
\begin{equation*}
% \begin{split}
            W_2(\mathcal{N} (\vmu_i, \mSigma_i), \mathcal{N} (\bm{0}, \mI)) 
            =     W_2(\mathcal{N} (\vmu_i, \mLambda_i), \mathcal{N} (\bm{0}, \mI)) \ \text{ for } i = 1, 2~,
% \end{split}
\end{equation*}
and the above claim is concluded.

Next, we vectorize the minimization problem in \eqref{eq:mindiagW2} as follows. We denote by $\R_+$ the set of positive real numbers.
Let $\vb$ be a general vector in $\R^K_+$, $\va'$ be a general vector in $\R^{\kappa}_+$ and $\va := (\va'; \bm{0}_{K-\kappa}) \in \R^K$. 
Given, the constraints on $\mSigma_1$ and $\mSigma_2$, we can parametrize 
the diagonal elements of $\mSigma_1^{\frac{1}{2}}$ and $\mSigma_2^{\frac{1}{2}}$ by $\va$ and $\vb$, that is, we set 
$\mSigma_1^{\frac{1}{2}} = $ diag($\va$) and $\mSigma_2^{\frac{1}{2}} = $ diag($\vb$). The objective function of \eqref{eq:mindiagW2} can then be written as 
\begin{equation*}\label{eq:min1W2}
    \eta \sqrt{\norm{\bm{\mu_1}}_2^2 + \norm{\va - \bm{1}_{K}}_2^2} + (1-\eta) \sqrt{\norm{\bm{\mu_2}}_2^2 + \norm{\vb - \bm{1}_{K}}_2^2}.
\end{equation*}
Combining this last expression and the same colinearity argument as in the proof of Proposition~\ref{prop:baryW1andKL} in \S\ref{subsec:proof} (supported by Fig.~\ref{fig:colinear}), \eqref{eq:mindiagW2} is equivalent to 
\begin{equation}\label{eq:min2W2}
% \begin{split}
    \min_{\substack{ \vmu_1, \vmu_2 \in \R^K, \\ \vb \in \R^K_{+}, \ \va' \in \R^{\kappa}_{+}, \ \va = (\va'; \bm{0}_{K-\kappa}), 
\\ 
(\vmu_1; \va), (\vmu_2; \vb), (\bm{0}_{K}; \bm{1}_{K}) ~\textup{are colinear}
\\ \& \norm{\vmu_1 - \vmu_2 }_2 =\epsilon}} 
\Biggl[ \eta \norm{(\vmu_1; \va) - (\bm{0}_{K}; \bm{1}_{K})}_2 
+ (1-\eta)\norm{(\vmu_2; \vb) - (\bm{0}_{K}; \bm{1}_{K})}_2 \Biggr].
% \end{split}
\end{equation}

{\bf Step III:} We solve \eqref{eq:min2W2}. By the colinearity constraint, we can write $(\bm{0}_{K}; \bm{1}_{K}) = u(\vmu_2; \vb) - (u-1) (\vmu_1; \va)$, where $u \in \R$. We thus obtain that
\begin{equation}\label{eq:simplified}
    \begin{split}
        (\vmu_2; \vb) - (\bm{0}_{K}; \bm{1}_{K}) & = (u-1) \left( (\vmu_1; \va) - (\vmu_2; \vb) \right)
    \\  (\vmu_1; \va) - (\bm{0}_{K}; \bm{1}_{K})  & = u \left( (\vmu_1; \va) - (\vmu_2; \vb) \right).
    \end{split}
\end{equation}
Furthermore, denoting the coordinates of $\va'$ and $\vb$ by $\{a_i\}_{i=1}^{\kappa}$ and $\{b_i\}_{i=1}^K$, we similarly obtain that
\begin{equation}\label{eq:simplified1}
    \begin{split}
      \bm{0}_{K} &= u\vmu_2 - (u-1)\vmu_1
    \\  1 &= ub_i - (u-1)a_i, \quad 1 \leq i \leq \kappa 
    \\  1 &= ub_i, \quad d+1 \leq i \leq K 
    \end{split}
\end{equation}
The last two of equations imply that 
$$ \sum_{i=1}^{\kappa} (a_i-b_i)^2 = \frac{\norm{\bm{1}_{\kappa}-\va'}_2^2}{u^2}$$ 
and 
$$\sum_{i=\kappa+1}^{K}b_i^2 = \frac{K-\kappa}{u^2}.$$ Combining \eqref{eq:commuteW2}, \eqref{eq:simplified} and the above two equations, we rewrite the objective function of \eqref{eq:min2W2} as follows:
\begin{align}
    & \bigl ( \eta |u| + |u-1| (1-\eta) \bigr)  \times \norm{(\vmu_1; \va) - (\vmu_2; \vb) }_2 \nonumber
    \\  
    = &\bigl ( \eta |u| + |u-1| (1-\eta) \bigr) \nonumber
     \times \sqrt{ \norm{\vmu_1 - \vmu_2}_2^2 + \sum_{i=1}^{\kappa} (a_i-b_i)^2 + \sum_{i=\kappa+1}^{K} b_i^2 } \nonumber
    \\ \geq & \left ( \eta |u| + |u-1| (1-\eta) \right ) \times \sqrt{ \epsilon^2 + \frac{\norm{\bm{1}_{\kappa}-\va'}_2^2}{u^2}  + \frac{K-\kappa}{u^2} } \nonumber
    \\  
    %\begin{split}
    \label{eq:min}
      = &\biggl\{(K-\kappa) \left ( (1-\eta) \left | \frac{u-1}{u} \right | + \eta \right )^2  
       + \epsilon^2 \bigl ( \eta |u| + |u-1| (1-\eta) \bigr)^2 
       + \norm{\bm{1}_{\kappa}-\va'}_2^2 \left ( (1-\eta) \left | \frac{u-1}{u} \right | + \eta \right )^2
        \biggr\}^{\!1/2},
    %\end{split}    
\end{align}

where equality is achieved if and only if $\norm{\vmu_1-\vmu_2}_2 =\epsilon$.
One can make the following two observations: $u=0$ does not yield a minimizer of \eqref{eq:min2W2}, and for any $u \neq 0 $, \eqref{eq:min} obtains its minimum at $\va' = \bm{1}_{\kappa}$. In view of these observations and the derivation above, we define 
\begin{equation}
% \begin{split}
        f(u) := (K-\kappa) \left ( (1-\eta) \left | \frac{u-1}{u} \right | + \eta \right )^2 
        + \epsilon^2 \left ( \eta |u| + |u-1| (1-\eta) \right )^2,
% \end{split}
\end{equation}
and note that \eqref{eq:min2W2} is equivalent to
\begin{equation}\label{eq:min3W2}
\min_{\substack{ u \neq 0}} 
\quad 
\sqrt{f(u)}.
\end{equation}

We rewrite $f(u)$ as 
\begin{equation*}\label{eq:fat1}
% \begin{split}
 f(u) = 
 \begin{cases}
%  \begin{split}
    \displaystyle (K-\kappa) \left(\frac{u-1}{u}(1-\eta) + \eta \right)^2 
     + \epsilon^2 \biggl( \eta u + (1-\eta)(u-1) \biggr)^2, 
%  \end{split} 
& u \geq 1 \\ 
%  \begin{split}
 \displaystyle (K-\kappa) \left(\frac{1-u}{u}(1-\eta) + \eta \right)^2 
 + \epsilon^2 \biggl( \eta u + (1-\eta)(1-u) \biggr)^2,
%  \end{split}
 & 1\geq u > 0 \\ 
% \begin{split}
\displaystyle (K-\kappa) \left(\frac{u-1}{u}(1-\eta) + \eta \right)^2 
+ \epsilon^2 \biggl( \eta u + (1-\eta)(u-1) \biggr)^2,
% \end{split}
&  0> u 
\end{cases}
% \end{split} 
\end{equation*}
We denote 
\begin{align*}
  r_1(u) :=(K-&\kappa) \left(\frac{u-1}{u}(1-\eta) + \eta \right)^2 
  + \epsilon^2 \biggl( \eta u + (1-\eta)(u-1) \biggr)^2
\end{align*}
and
\begin{align*}
    r_2(u):=(K-&\kappa) \left(\frac{1-u}{u}(1-\eta) + \eta \right)^2 
    + \epsilon^2 \biggl( \eta u + (1-\eta)(1-u) \biggr)^2.  
\end{align*}
Their derivatives are $$r_1'(u) = \frac{2}{u^3} \left( u-(1-\eta) \right) \left( \epsilon^2 u^3 +(K-\kappa)(1-\eta) \right)$$ 
and 
\begin{equation*}
    r_2'(u) = \frac{2}{u^3} \biggl( (2\eta-1)u + (1-\eta) \biggr) \times \left( \epsilon^2 (2\eta-1)u^3 - (K-\kappa)(1-\eta) \right).
\end{equation*}
These expressions for $r_1'$ and $r_2'$ imply that the critical points for $r_1$ are 
$$u_{r_1}^{(1)} = 1-\eta \ \text{ and } \ u_{r_1}^{(2)} = - \left( \frac{(K-\kappa)(1-\eta)}{\epsilon^2} \right)^{\frac{1}{3}}$$ 
and the critical points for $r_2$ are 
$$u_{r_2}^{(1)} = -\left( \frac{1-\eta}{2\eta - 1} \right) \ \text{ and } \ u_{r_2}^{(2)} =  \left( \frac{(K-\kappa)(1-\eta)}{\epsilon^2(2\eta-1)} \right)^{\frac{1}{3}}.$$

We note that $r_1$ is increasing on $(u_{r_1}^{(2)}, 0) \cup (u_{r_1}^{(1)}, \infty)$ and decreasing on $(-\infty, u_{r_1}^{(2)}) \cup (0, u_{r_1}^{(1)})$. On the other hand, $r_2$ is increasing on $(u_{r_2}^{(1)}, 0) \cup (u_{r_2}^{(2)}, \infty)$ and decreasing on $(-\infty, u_{r_2}^{(1)}) \cup (0, u_{r_2}^{(2)})$. Since $\eta> \eta^{\star} = \displaystyle\frac{K-\kappa+\epsilon^2}{K-\kappa+2\epsilon^2}$, $u_{r_2}^{(2)} \in (0, 1)$. The derivative of $f$ with respect to $u$ is
\begin{equation*}
f'_{u}(u) =
 \begin{cases} 
r_1'(u), & u>0 \\ 
r_2'(u), & 1>u>0 \\ 
r_1'(u), & 0>u. 
\end{cases}
\end{equation*}
So $f(\bm{\cdot})$ is increasing on $(u_{r_1}^{(2)}, 0) \cup (u_{r_2}^{(2)}, \infty)$ and decreasing on $(-\infty, u_{r_1}^{(2)}) \cup (0, u_{r_2}^{(2)})$. The values of $f$ at $u_{r_2}^{(2)}$ and $u_{r_1}^{(2)}$ are 
\begin{equation*}
    \begin{split}
        &\begin{split}
        f(u_{r_2}^{(2)}) =        \Biggl( \biggl( &\frac{(K-\kappa)(1-\eta)(2\eta-1)^2}{\epsilon^2} \biggr)^{\frac{1}{3}} + (1-\eta) \Biggr)^2 
                \times \left( (K-\kappa)^{\frac{1}{3}} \left( \frac{\epsilon^2 (2\eta-1)}{(1-\eta)} \right)^{\frac{2}{3}} + \epsilon^2 \right),    
        \end{split}
    \\&\begin{split}
      f(u_{r_1}^{(2)}) =           \Biggl( \biggl( &\frac{(K-\kappa)(1-\eta)}{\epsilon^2} \biggr)^{\frac{1}{3}} + (1-\eta) \Biggr)^2 
                \times \left( (K-\kappa)^{\frac{1}{3}} \left( \frac{\epsilon^2}{(1-\eta)} \right)^{\frac{2}{3}} + \epsilon^2 \right). 
    \end{split}
    \end{split}
\end{equation*}
Consequently, the minimum of $f$ is obtained at $u^{\star} := u_{r_2}^{(2)}$. By \eqref{eq:simplified} and \eqref{eq:simplified1},  the means $\vmu_1$, $\vmu_2$ and the covariance matrices $\mSigma_1$, $\mSigma_2$ satisfy:
$\bm{0}_K = u^{\star}\bm{\mu_2} + (1-u^{\star})\bm{\mu_1} $, $\mSigma_1=\textup{diag}(\bm{1}_\kappa; \bm{0}_{K-\kappa})$ and $\mSigma_2=\textup{diag}(\bm{1}_\kappa; {(u^{\star})}^{-2}\bm{1}_{K-\kappa})$. Moreover, the norms of $\vmu_1$ and $\vmu_2$ can be computed from \eqref{eq:simplified1} as $u^{\star}\epsilon$ and $(1-u^{\star})\epsilon$, respectively.

\subsection{Proof of Proposition~\ref{prop:baryKL}}\label{subsec:proofbaryKL}

Notice that since $\mSigma_0 \in {\cal S}_{++}^{K}$, det$(\mSigma_0)>0$. On the other hand, since $\mSigma_1 \in {\cal S}_{+}^{K}$ with rank$(\mSigma_1) = \kappa < K$, det$(\mSigma_1)=0$. Therefore, 
\begin{equation*}
    \log \dfrac{ \textup{det}(\mSigma_0)}{\textup{det}(\mSigma_1)} = \log \textup{det}(\mSigma_0) - \log \textup{det}(\mSigma_1)  = \infty.
\end{equation*}
This and \eqref{eq:closedKL} imply that
$  KL(\mathcal{N}(\vmu_1, \mSigma_1) || \mathcal{N}(\vmu_0, \mSigma_0)) = \infty$.

\section{ADDITIONAL DETAILS ON THE BENCHMARK METHODS}
\label{sec:benchmark}
We overview the benchmark methods compared with MAW, where we present them according to alphabetical order of names. 
We will include all tested codes in a supplemental webpage.

For completeness, we mention the following links (or papers with links) we used for the different codes. For DSEBMs and DAGMM we used the codes of %linked in
\citep{golan2018deep}. For LOF, OCSVM and IF we used the scikit-learn \citep{sklearn_api} packages for novelty detection.
For OCGAN we used its TensorFlow implementation from \url{https://pypi.org/project/ocgan}. 
For RSRAE, we adapted the code %linked in
of \citep{lai2020robust} to novelty detection. 

All experiments were executed on a Linux machine with 64GB RAM and four GTX1080Ti GPUs.

We remark that for the neural networks based methods (DAGMM, DSEBMs, OCGAN and RSRAE), we followed similar implementation details as the one described in \S\ref{sec:MAW_implement} for MAW.

\noindent
\textbf{Deep Autoencoding GMM (DAGMM) \citep{zong2018deep}:} This method uses a deep autoencoder model. It optimizes an end-to-end structure that contains both an autoencoder and an estimator for a GMM. Anomalies are detected using this GMM.
We remark that this mixture model is proposed for the inliers.
An improved version of DAGMM was recently proposed in \citep{fan2020correlation}. 

\noindent
\textbf{Deep Structured Energy-Based Models (DSEBMs) \citep{zhai2016deep}:} Its decision is based on an energy function which is the negative log probability that a sample follows the data distribution. An autoencoder is used for the energy-based model 
in order to avoid the need of complex sampling. 

\noindent
\textbf{Isolation Forest (IF) \citep{liu2008isolation}:} It iteratively constructs special binary trees for the training set and identifies anomalies in the test set as the ones with shortest average path lengths.

\noindent
\textbf{Local Outlier Factor (LOF) \citep{breunig2000lof}:} It measures the isolation of a data point from its surrounding neighbors by estimating the local density of this point using its $k$ nearest neighbors. In novelty detection, it identifies novelties according to low density regions learned from the training data.

\noindent
\textbf{One-class Novelty Detection Using GANs (OCGAN) \citep{perera2019ocgan}:} It is composed of four NNs: a denoising autoencoder, two adversarial discriminators, and a classifier. It adversarially encourages the autoencoder to learn only the inlier features. 

\noindent
\textbf{One-Class SVM (OCSVM) \citep{heller2003one}:} It estimates the margin of the training set and uses it as the decision boundary for the test set. It commonly utilizes a radial basis function kernel.

\noindent
\textbf{Robust Subspace Recovery Autoencoder (RSRAE) \citep{lai2020robust}:} It uses an autoencoder with a linear RSR layer and an $\ell_{2,1}$-based penalty. The RSR layer extracts features of inliers in the latent code while helping to reject outliers. The instances with higher reconstruction errors are viewed as outliers. RSRAE trains a model using the training data. We then apply this model for detecting novelties in the test data.

\section{ADDITIONAL DETAILS ON THE DIFFERENT DATASETS} \label{sec:datasets}

Below we provide additional details on the six  datasets used in our experiments. We remark that each dataset contains several clusters (3 for COVID-19, 10 for CIFAR-10, 11 largest ones for Caltech101, 10 for Fashion MNIST, 2 for KDDCUP-99 and 5 for Reuters-21578, ). Table~\ref{tab:numdata} lists for each dataset (for both training and testing) the data types, numbers of clusters, dimensions, numbers of instances and numbers of inliers and outliers.

\begin{table*}[ht]
  \caption{Summary of properties of the datasets.}
%  , numbers of clusters, numbers of instances, dimensions and the numbers of inliers and outliers for training and testing. 
%  }
  \centering
  \resizebox{0.9\textwidth}{!}{%
  
 \begin{tabular}{lccccccccccc}
    \toprule
    \multicolumn{1}{}{} & \multicolumn{5}{c}{ Dataset information} &  
    \multicolumn{1}{}{} & \multicolumn{2}{c}{Training} & \multicolumn{1}{}{} & \multicolumn{2}{c}{Testing} \\
    \cmidrule(r){2-6}
    \cmidrule(r){8-9}
    \cmidrule(r){11-12}
    Datasets     & {Type}  & {$\#$Clusters}  & {Dimension} & {$\#$Instances}   &   &   & $\#$Inliers  & $\#$Outliers  &  & $\#$Inliers  & $\#$Outliers 
    % \\ &   &  &  &  &  &
    % & ($N$)& ($N \times c$ ) & & ($N_{\textup{test}}$) & ($N_{\textup{test}} \times c_{\textup{test}}$)
    \\
    \midrule
    COVID-19 (Radiography) & Image  & {3} & $64 \times 64 \times 3$  &  {15,161}  &   &    & 160& 160 $\times c$ & & 60 & 60 $\times c_{\textup{test}}$ \\
    CIFAR-10  & Image  & {10} & $32 \times 32 \times 3$ &  {60,000}   &   &      & 450  & 450 $\times c$   &  & 150 & 150 $\times c_{\textup{test}}$ \\  
    Caltech101  & Image  & {11} & $32 \times 32 \times 3$ &  {9,146} &   &    & 100  & 100 $\times c$   &  & 100 & 100 $\times c_{\textup{test}}$ \\    
    Fashion MNIST  &  Image  & {10} & $28 \times 28 \times 1$ &   {70,000}  &   &  & 300 & 300 $\times c$ && 60 & 60 $\times c_{\textup{test}}$\\
    KDDCUP-99   &  Feature & {2} & $120$ &  {494,021}  &   &    & 6000 & 6000 $\times c$  & & 1200 & 1200 $\times c_{\textup{test}}$  \\ 
    Reuters-21578  & Feature & {5} & $26,147$ &  {21,578}  &   &   & 350 & 350 $\times c$  & & 140  & 140 $\times c_{\textup{test}}$ 
    \\
    \bottomrule
  \end{tabular}
  }
  \label{tab:numdata}
\end{table*}

\noindent
\textbf{COVID-19 (Radiography) \citep{chowdhury2020can}:} It contains chest X-ray images (RGB) labeled according to these  categories: COVID-19 positive, normal and bacterial Pneumonia cases. We resize the images to size $64 \times 64$ and rescale the pixel intensities to lie in $[-1, 1]$. It is publicly available in \url{https://www.kaggle.com/tawsifurrahman/covid19-radiography-database}.

\noindent
\textbf{CIFAR-10 \citep{krizhevsky2009learning}:} It contains 10 categories of $32 \times 32$ RGB images of transportation vehicles and animals.   We rescale the pixel intensities to lie in $[0, 1]$. The dataset is publicly available in \url{https://www.cs.toronto.edu/~kriz/cifar.html}.

\noindent
\textbf{Caltech101 \citep{fei2004learning}:} It contains RGB images of objects from 101 categories with identifying labels.
Following \citet{lai2020robust}, we use the largest 11 classes and preprocess their images to have size $32 \times 32$ and rescale the pixel intensities to lie in $[-1,1]$. 
It is publicly available in \url{http://www.vision.caltech.edu/Image_Datasets/Caltech101}. 

\noindent
\textbf{Fashion MNIST \citep{xiao2017fashion}:} It is an image dataset containing 10 categories of $28 \times 28$ grayscale images of clothing and accessories items. We rescaled the pixel intensities to lie in $[-1,1]$. We obtained the dataset from the Keras dataset library \url{https://keras.io/api/datasets/fashion_mnist}.

\noindent
\textbf{KDDCUP-99 \citep{dua2017uci}:} It is a classic dataset for intrusion detection. It contains feature vectors of connections between internet protocols and a binary label for each feature vector identifying normal vs.~abnormal ones. The abnormal ones are associated with an ``attack'' or ``intrusion''. The dataset is publicly available in \url{http://kdd.ics.uci.edu/databases/kddcup99/kddcup99.html}.

\noindent
\textbf{Reuters-21578 \citep{lewis1997reuters}:} It contains 21,578 documents with 90 text categories having multi-labels. Following \citet{lai2020robust}, we consider the five largest classes with single labels. We utilize the scikit-learn packages: TFIDF and HashingVectorizer \citep{rajaraman2011mining} to preprocess the documents into 26,147 dimensional vectors. It is publicly available in \url{https://archive.ics.uci.edu/ml/datasets/reuters-21578+text+categorization+collection}.

According to the above description, the numbers of clusters of these datasets are 3, 10, 11, 10, 2 and 5,  respectively.
We remark that COVID-19, CIFAR-10,  Caltech101, Fashion MNIST and Reuters-21578 separate between training and testing data points. For KDDCUP-99, we randomly split it into training and testing datasets of equal sizes.

\section{NUMERICAL RESULTS OF THE EXPERIMENTS}\label{sec:numerical}

We present as tables the numerical values depicted in Figs.~\ref{fig:res} and~\ref{fig:variations}. 
Tables~\ref{tab:auc_covid}-\ref{tab:ap_reuters} report the averaged AUC and AP scores with training outliers/inliers ratio $c \in \{0.1, 0.2, 0.3, 0.4, 0.5\}$ that were depicted in Fig.~\ref{fig:res}. Each table describes one of the averaged scores (AUC or AP) for one of the six datasets (COVID-19, CIFAR-10, Caltech101, Fashion MNIST, KDDCUP-99 and Reuters-21578) and also indicates the standard deviation of each value. The outperforming methods are marked in bold.

Tables~\ref{tab:var_kdd_auc}-\ref{tab:var_cov_ap} record the averaged AUC and AP scores with training outliers/inliers ratio $c \in \{ 0.1, 0.2, 0.3, 0.4, 0.5\}$ that were depicted in Fig.~\ref{fig:variations}. Each table describes one of the averaged scores (AUC or AP) for one of either KDDCUP-99 or COVID-19 and also indicates the standard deviation of each value. The outperforming methods are boldfaced.

\begin{table}[h!]
  \caption{AUC scores of COVID-19.}
  \label{tab:auc_covid}
  \centering
  \resizebox{0.67\columnwidth}{!}{%
  \begin{tabular}{lccccc}
    \toprule
    \multicolumn{1}{}{} & \multicolumn{5}{c}{Training ratio of outliers per inliers, $c$} \\
    \cmidrule(r){2-6}
    Methods &  0.1 & 0.2 & 0.3 & 0.4 & 0.5\\
    \midrule
    MAW     &  \textbf{0.652} $\pm$ 0.021 & \textbf{0.609} $\pm$ 0.018 & \textbf{0.576} $\pm$ 0.019 & 0.531 $\pm$ 0.020 & 0.504 $\pm$ 0.010\\
    DAGMM   &   0.527$\pm$ 0.068 &  0.545$\pm$ 0.051 & 0.518 $\pm$ 0.062 & 0.504 $\pm$ 0.060 &  0.503 $\pm$ 0.057\\
    DSEBMs  &   0.451$\pm$ 0.000 &  0.451$\pm$ 0.000 & 0.451 $\pm$ 0.000& 0.451 $\pm$ 0.000 & 0.451 $\pm$ 0.000\\
    IF      &  	0.574   & 	0.541     & 0.515	& 0.493 & 0.469	\\
    LOF     &  	0.642   & 	 0.588    & 0.542	& \textbf{0.536} & \textbf{0.519}\\
    OCGAN   &   0.472$\pm$ 0.000 &  0.472$\pm$ 0.000 & 0.465 $\pm$ 0.000& 0.445 $\pm$ 0.000 & 0.431 $\pm$ 0.000 \\
    OCSVM   &  	0.528   & 	0.528     & 0.528	&  0.535 &	0.521\\
    RSRAE   &   0.535$\pm$ 0.031 &  0.507$\pm$ 0.028 & 0.456 $\pm$ 0.023& 0.434 $\pm$ 0.018 &  0.407 $\pm$ 0.011\\
    \bottomrule
  \end{tabular}}
\end{table}

\begin{table}[h!]
  \caption{AP scores of COVID-19.}
  \label{tab:ap_covid}
  \centering
  \resizebox{0.67\columnwidth}{!}{%
  \begin{tabular}{lccccc}
    \toprule
    \multicolumn{1}{}{} & \multicolumn{5}{c}{Training ratio of outliers per inliers,  $c$} \\
    \cmidrule(r){2-6}
    Methods &  0.1 & 0.2 & 0.3 & 0.4 & 0.5\\
    \midrule
    MAW    & 0.459 $\pm$ 0.014 & \textbf{0.442} $\pm$ 0.011& \textbf{0.424} $\pm$ 0.018 & 0.368 $\pm$ 0.015 & 0.353 $\pm$ 0.013\\
    DAGMM   &  0.354$\pm$ 0.053& 0.390 $\pm$ 0.057& 0.316 $\pm$ 0.052 & 0.357 $\pm$ 0.050  & 0.348 $\pm$ 0.047\\
    DSEBMs  &  0.372$\pm$ 0.000& 0.375 $\pm$ 0.000& 0.364 $\pm$ 0.000 & 0.360 $\pm$  0.000 & 0.358 $\pm$ 0.000 \\    
    IF      & 	0.425& 0.404	& 0.392	& 0.373 & 0.363	\\
    LOF     & 	\textbf{0.463}& 0.422	& 0.402	& \textbf{0.374} & \textbf{0.371}	\\  
    OCGAN   &  0.381$\pm$ 0.000& 0.381 $\pm$ 0.000& 0.381 $\pm$ 0.000 & 0.373 $\pm$  0.000 & 0.350 $\pm$ 0.000 \\  
    OCSVM   & 	0.315& 0.315	& 0.315	& 0.372 & 0.365	\\
    RSRAE   &  0.388$\pm$ 0.018& 0.377 $\pm$ 0.016& 0.355 $\pm$ 0.011 & 0.352 $\pm$  0.010 & 0.340 $\pm$ 0.009\\
    \bottomrule
  \end{tabular}}
\end{table}

\begin{table}[h!]
  \caption{AUC scores of CIFAR-10.}
  \label{tab:auc_cifar}
  \centering
  \resizebox{0.67\columnwidth}{!}{
  \begin{tabular}{lccccc}
    \toprule
    \multicolumn{1}{}{} &\multicolumn{5}{c}{Training ratio of outliers per inliers, $c$} \\
    \cmidrule(r){2-6}
    Methods & 0.1 & 0.2 & 0.3  & 0.4 & 0.5\\
    \midrule
    MAW     & 0.621 $\pm$ 0.013 & \textbf{0.609} $\pm$ 0.014& \textbf{0.607} $\pm$ 0.012 & 0.600 $\pm$ 0.010& \textbf{0.595} $\pm$	0.013\\
    LOF     & 0.582	& 0.574	&  0.559 & 0.551  &	0.539\\
    OCSVM   & 0.595	& 0.587	& 0.580	& 0.564  &	0.570\\
    IF      & 0.603	& 0.586	& 0.596  & 0.581  &	0.569\\
    RSRAE   & \textbf{0.638} $\pm$ 0.010 & 0.607 $\pm$ 0.017 & 0.599 $\pm$ 0.023& \textbf{0.610} $\pm$ 0.025& 0.589 $\pm$ 0.023\\
    DSEBMs  & 0.586 $\pm$ 0.006 & 0.584 $\pm$0.006 & 0.580 $\pm$ 0.004&  0.576$\pm$ 0.006& 0.556 $\pm$0.006\\
    OCGAN   & 0.501 $\pm$0 & 0.501 $\pm$0 & 0.499 $\pm$0 & 0.487 $\pm$ 0& 0.476 $\pm$0\\
    DAGMM   & 0.574 $\pm$ 0.030& 0.557 $\pm$ 0.035& 0.541 $\pm$0.037 & 0.510 $\pm$ 0.0331& 0.545  $\pm$ 0.037\\
    \bottomrule
  \end{tabular}}
\end{table}

\begin{table}[h!]
  \caption{AP scores of CIFAR-10.}
  \label{tab:ap_cifar}
  \centering
  \resizebox{0.67\columnwidth}{!}{
  \begin{tabular}{lccccc}
    \toprule
    \multicolumn{1}{}{} &\multicolumn{5}{c}{Training ratio of outliers per inliers, $c$} \\
    \cmidrule(r){2-6}
    Methods & 0.1 & 0.2 & 0.3  & 0.4 & 0.5\\
    \midrule
    MAW    & 0.427 $\pm$ 0.010& \textbf{0.419} $\pm$0.012 & 0.414 $\pm$ 0.011&  \textbf{0.400}$\pm$0.009 & \textbf{0.411} $\pm$0.011	\\
    LOF     & 0.395	& 	0.036&  0.377 & 0.374  & 0.371	\\
    OCSVM   & 0.408	& 0.400	& 0.393	& 0.378  &	0.385\\
    IF      & 0.416	& 0.395	&  0.403 & 0.389  &	0.373\\
    RSRAE   & \textbf{0.434} $\pm$ 0.011& 0.412 $\pm$0.020 & \textbf{0.417} $\pm$ 0.022& 0.391 $\pm$ 0.019& 0.400 $\pm$ 0.014\\
    DSEBMs  & 0.391 $\pm$0.008 & 0.388 $\pm$ 0.008& 0.386 $\pm$0.004 & 0.382 $\pm$ 0.006&  0.379$\pm$0.003\\
    OCGAN   & 0.342 $\pm$ 0&  0.340$\pm$0 & 0.339 $\pm$0 & 0.337 $\pm$ 0 & 0.335$\pm$0\\
    DAGMM   & 0.378 $\pm$0.049 & 0.369 $\pm$0.041 & 0.355 $\pm$ 0.030& 0.308 $\pm$ 0.026& 0.352 $\pm$ 0.047\\
    \bottomrule
  \end{tabular}}
\end{table}

\begin{table}[h!]
  \caption{AUC scores of Caltech101.}
  \label{tab:auc_caltech}
  \centering
  \resizebox{0.67\columnwidth}{!}{
  \begin{tabular}{lccccc}
    \toprule
    \multicolumn{1}{}{} & \multicolumn{5}{c}{Training ratio of outliers per inliers, $c$} \\
    \cmidrule(r){2-6}
    Methods &  0.1 & 0.2 & 0.3 & 0.4 & 0.5\\
    \midrule
    MAW     &  \textbf{0.801} $\pm$ 0.017 & \textbf{0.760} $\pm$  0.028& \textbf{0.700} $\pm$ 0.038 & \textbf{0.608} $\pm$ 0.031 & \textbf{0.570} $\pm$ 0.021 \\
    DAGMM   &  0.684 $\pm$ 0.100 &  0.588 $\pm$ 0.115 &  0.500$\pm$ 0.100 & 0.509 $\pm$ 0.101 & 0.514 $\pm$ 0.095\\
    DSEBMs  &  0.536 $\pm$ 0.011 &  0.612$\pm$ 0.025 & 0.577 $\pm$ 0.030 & 0.564 $\pm$ 0.021 & 0.536 $\pm$ 0.021\\
    IF      &  0.755	   & 	0.694     & 0.626	& 0.575 &	0.540\\
    LOF     &  0.674	   & 	 0.593    & 0.495	& 0.436 &	0.411\\
    OCGAN   &  0.494 $\pm$ 0.000 &  0.494$\pm$ 0.000 &  0.494$\pm$ 0.000 & 0.500 $\pm$ 0.000 & 0.500 $\pm$ 0.000\\
    OCSVM   &  0.682	   & 	0.618     & 0.577	& 0.538 &	0.516\\
    RSRAE   &  0.774 $\pm$ 0.027 & 0.722 $\pm$ 0.041 & 0.664 $\pm$ 0.082  & 0.579 $\pm$ 0.047 & 0.568 $\pm$ 0.036 \\
    \bottomrule
  \end{tabular}}
\end{table}

\begin{table}[h!]
  \caption{AP scores of Caltech101.}
  \label{tab:ap_caltech}
  \centering
  \resizebox{0.67\columnwidth}{!}{
  \begin{tabular}{lccccc}
    \toprule
    \multicolumn{1}{}{} & \multicolumn{5}{c}{Training ratio of outliers per inliers, $c$} \\
    \cmidrule(r){2-6}
    Methods &  0.1 & 0.2 & 0.3 & 0.4 & 0.5\\
    \midrule
    MAW     &  \textbf{0.634} $\pm$ 0.027 & \textbf{0.572} $\pm$ 0.039 & \textbf{0.531} $\pm$ 0.064 & 0.412 $\pm$ 0.029 & 0.414 $\pm$ 0.021\\
    DAGMM   &   0.574$\pm$ 0.088 & 0.422 $\pm$ 0.112 & 0.308 $\pm$ 0.102 & 0.351 $\pm$ 0.074 & 0.363 $\pm$ 0.076\\
    DSEBMs  &   0.385$\pm$ 0.003 &  0.472$\pm$ 0.051 &  0.398$\pm$0.019 & 0.383 $\pm$ 0.023 & 0.365 $\pm$ 0.028\\
    IF      &  	0.545   & 	0.486     & 0.430	& 0.304 & 0.371	\\
    LOF     &  	0.460   & 	0.400     & 0.337	& 0.304 & 0.290	\\
    OCGAN   &   0.362$\pm$ 0.000 &  0.362$\pm$ 0.000 & 0.362 $\pm$ 0.000 & 0.362 $\pm$ 0.000 & 0.362 $\pm$ 0.000\\
    OCSVM   &  	0.472   & 	0.419     & 0.380	& 0.352 & 0.339	\\
    RSRAE   &   0.595$\pm$ 0.038 & 0.551 $\pm$ 0.045 & 0.495 $\pm$0.073 & \textbf{0.425} $\pm$ 0.040 & \textbf{0.443} $\pm$ 0.027\\
    \bottomrule
  \end{tabular}}
\end{table}

\begin{table}[h!]
  \caption{AUC scores of Fashion MNIST}
  \label{tab:auc_fashion}
  \centering
  \resizebox{0.67\columnwidth}{!}{
  \begin{tabular}{lccccc}
    \toprule
    \multicolumn{1}{}{} & \multicolumn{5}{c}{Training ratio of outliers per inliers, $c$} \\
    \cmidrule(r){2-6}
    Methods &  0.1 & 0.2 & 0.3 & 0.4 & 0.5\\
    \midrule
    MAW    & \textbf{0.897} $\pm$ 0.013 & \textbf{0.879} $\pm$ 0.011 & \textbf{0.852} $\pm$ 0.022 & 0.830 $\pm$ 0.017 & 0.801  $\pm$	0.016\\
    DAGMM   & 0.607 $\pm$ 0.093 & 0.376 $\pm$ 0.070 & 0.427  $\pm$ 0.090 & 0.401 $\pm$ 0.078& 0.411 $\pm$ 0.081\\
    DSEBMs  & 0.730 $\pm$ 0.092 & 0.729 $\pm$ 0.105 & 0.739 $\pm$ 0.086 & 0.723 $\pm$ 0.106& 0.687  $\pm$ 0.096\\
    IF      & 0.893	& 0.875	& 0.843	& \textbf{0.834} & \textbf{0.827}    \\
    LOF     &  0.569	& 0.507 	& 0.476   & 0.468 	& 0.458	\\
    OCGAN   & 0.542 $\pm$ 0.006 & 0.538 $\pm$ 0.004 & 0.544 $\pm$ 0.014 & 0.531 $\pm$ 0.003 & 0.525 $\pm$ 0.004\\
    OCSVM   &  0.895	& 0.874	& 0.848	& 0.831	& 0.814\\
    RSRAE   & 0.860 $\pm$ 0.022 & 0.848 $\pm$ 0.022 & 0.829 $\pm$ 0.042 & 0.831 $\pm$ 0.028 &0.808  $\pm$ 0.028\\

    \bottomrule
  \end{tabular}}
\end{table}

\begin{table}[h!]
  \caption{AP scores of Fashion MNIST}
  \label{tab:ap_fashion}
  \centering
  \resizebox{0.67\columnwidth}{!}{
  \begin{tabular}{lccccc}
    \toprule
    \multicolumn{1}{}{} & \multicolumn{5}{c}{Training ratio of outliers per inliers, $c$} \\
    \cmidrule(r){2-6}
    Methods &  0.1 & 0.2 & 0.3 & 0.4 & 0.5\\
    \midrule
    MAW    & \textbf{0.788}  $\pm$0.013 & 0.754 $\pm$ 0.014&  0.723$\pm$0.029  & 0.686 $\pm$ 0.025& 0.672 $\pm$0.021\\
    DAGMM   & 0.482 $\pm$0.051 & 0.303 $\pm$0.057 & 0.334 $\pm$0.113  & 0.318 $\pm$0.056 & 0.330 $\pm$ 0.038\\
    DSEBMs  & 0.600 $\pm$ 0.045&  0.609$\pm$ 0.120&  0.613$\pm$0.089  & 0.605 $\pm$0.086 & 0.565 $\pm$ 0.072\\
    IF      & 0.768	& 0.724	& 0.693	& 0.665  & 0.642    \\
    LOF     & 0.382	& 0.331	& 0.308  & 0.301 & 0.294	\\
    OCGAN   & 0.504 $\pm$ 0.002& 0.503 $\pm$ 0.003& 0.500 $\pm$ 0.059  & 0.495  $\pm$ 0.001 & 0.493 $\pm$ 0.001\\ 
    OCSVM   & 0.801	& \textbf{0.768}	& \textbf{0.735}	& \textbf{0.696}	& 0.664	\\
    RSRAE   & 0.749 $\pm$ 0.029& 0.736 $\pm$ 0.032& 0.716 $\pm$ 0.048  & 0.683  $\pm$ 0.036& \textbf{0.680} $\pm$ 0.042\\
    \bottomrule
  \end{tabular}}
\end{table}

\begin{table}[h!]
  \caption{AUC scores of KDDCUP-99.}
  \label{tab:auc_kdd}
  \centering
  \resizebox{0.67\columnwidth}{!}{
  \begin{tabular}{lccccc}
    \toprule
    \multicolumn{1}{}{} & \multicolumn{5}{c}{Training ratio of outliers per inliers, $c$} \\
    \cmidrule(r){2-6}
    Methods &  0.1 & 0.2 & 0.3 & 0.4 & 0.5\\
    \midrule
    MAW     &  \textbf{0.945} $\pm$ 0.028 & \textbf{0.906} $\pm$ 0.018 & \textbf{0.832} $\pm$ 0.016 & \textbf{0.775} $\pm$ 0.023 & \textbf{0.731} $\pm$ 0.017\\
    DAGMM   &  0.614 $\pm$ 0.083 & 0.660 $\pm$ 0.109 & 0.584 $\pm$ 0.133& 0.457 $\pm$ 0.099 & 0.521 $\pm$ 0.089\\    
    DSEBMs  &  0.514$\pm$  0.000 & 0.499 $\pm$ 0.000 & 0.497 $\pm$ 0.000 & 0.496 $\pm$  0.000& 0.496 $\pm$ 0.000\\
    IF      &  0.811	   & 0.850	     & 0.807 & 0.750 & 0.706	\\
    LOF     &  0.480	   & 0.527	     & 0.516 & 0.527 & 0.530	\\
    OCGAN   &  0.651 $\pm$ 0.157 & 0.552 $\pm$ 0.157 & 0.617 $\pm$ 0.191 & 0.517 $\pm$ 0.146 & 0.628 $\pm$ 0.155 \\
    OCSVM   &  0.502	   & 0.568	     & 0.567	& 0.555 & 0.534	\\
    RSRAE   &  0.815 $\pm$ 0.031 & 0.839 $\pm$ 0.059 & 0.774 $\pm$ 0.086 & 0.735 $\pm$ 0.066 & 0.710 $\pm$ 0.056\\
    \bottomrule
  \end{tabular}}
\end{table}

\begin{table}[h!]
  \caption{AP scores of KDDCUP-99.}
  \label{tab:ap_kdd}
  \centering
  \resizebox{0.67\columnwidth}{!}{
  \begin{tabular}{lccccc}
    \toprule
    \multicolumn{1}{}{} & \multicolumn{5}{c}{Training ratio of outliers per inliers, $c$} \\
    \cmidrule(r){2-6}
    Methods &  0.1 & 0.2 & 0.3 & 0.4 & 0.5\\
    \midrule
    MAW    & \textbf{0.765}$\pm$ 0.025& \textbf{0.732} $\pm$ 0.015& \textbf{0.647} $\pm$ 0.012& \textbf{0.594} $\pm$ 0.014 & 0.556 $\pm$ 0.014\\
    DAGMM   & 0.446 $\pm$ 0.047& 0.506 $\pm$ 0.064& 0.459 $\pm$ 0.087& 0.373 $\pm$ 0.109 & 0.464 $\pm$ 0.998\\
    DSEBMs  & 0.450 $\pm$ 0.000& 0.447 $\pm$ 0.000& 0.446 $\pm$ 0.000& 0.444 $\pm$ 0.000 & 0.444 $\pm$ 0.000 \\
    IF      & 0.636	& 0.6331& 0.562	& 0.493 & 0.457	\\
    LOF     & 0.391	& 0.407	& 0.392	& 0.394 & 0.391	\\
    OCGAN   & 0.582 $\pm$ 0.132& 0.472 $\pm$ 0.163& 0.525 $\pm$ 0.133& 0.418 $\pm$ 0.136  & 0.535 $\pm$ 0.133\\
    OCSVM   & 0.543	& 0.598	& 0.595	& 0.438 & 0.426	\\
    RSRAE   & 0.704 $\pm$ 0.048& 0.698 $\pm$ 0.050 & 0.606 $\pm$ 0.065& 0.584 $\pm$ 0.034 & \textbf{0.574} $\pm$ 0.046\\
    \bottomrule
  \end{tabular}}
\end{table}

\begin{table}[h!]
  \caption{AUC scores of Reuters-21578.}
  \label{tab:auc_reuters}
  \centering
  \resizebox{0.67\columnwidth}{!}{
  \begin{tabular}{lccccc}
    \toprule
    \multicolumn{1}{}{} & \multicolumn{5}{c}{Training ratio of outliers per inliers, $c$} \\
    \cmidrule(r){2-6}
    Methods &  0.1 & 0.2 & 0.3 & 0.4 & 0.5\\
    \midrule
    MAW     &  0.885 $\pm$ 0.028 & \textbf{0.830} $\pm$ 0.013 & 0.770  $\pm$ 0.017 & \textbf{0.700} $\pm$ 0.002 & \textbf{0.648} $\pm$ 0.016\\
    DAGMM   &  0.500 $\pm$  0.000 & 0.511 $\pm$ 0.027 & 0.566 $\pm$ 0.110 & 0.559 $\pm$ 0.087 & 0.570 $\pm$ 0.091\\
    DSEBMs  &  \textbf{0.887} $\pm$  0.012& 0.825 $\pm$  0.012& \textbf{0.790} $\pm$ 0.015 & 0.690 $\pm$ 0.002 & 0.648 $\pm$ 0.010\\
    IF      &  	0.544   & 	 0.535    & 0.520	& 0.453 & 0.452	\\
    LOF     &  	0.757   & 	0.612     & 0.579	& 0.631 & 0.616\\
    OCGAN   &  0.648 $\pm$  0.127 & 0.477 $\pm$ 0.129 & 0.498 $\pm$ 0.140 & 0.519 $\pm$ 0.132 & 0.502 $\pm$ 0.099\\
    OCSVM   &  	0.882   & 	 0.817    & 0.785	& 0.673 &	0.640\\
    RSRAE   &   0.786 $\pm$ 0.042 & 0.755 $\pm$ 0.034 & 0.716 $\pm$ 0.033 & 0.605 $\pm$ 0.001 & 0.494 $\pm$ 0.004 \\
    \bottomrule
  \end{tabular}}
\end{table}

\begin{table}[h!]
  \caption{AP scores of Reuters-21578.}
  \label{tab:ap_reuters}
  \centering
  \resizebox{0.67\columnwidth}{!}{
  \begin{tabular}{lccccc}
    \toprule
    \multicolumn{1}{}{} & \multicolumn{5}{c}{Training ratio of outliers per inliers, $c$} \\
    \cmidrule(r){2-6}
    Methods &  0.1 & 0.2 & 0.3 & 0.4 & 0.5\\
    \midrule
    MAW     &  0.755 $\pm$ 0.041 & 0.677 $\pm$ 0.026 & 0.627 $\pm$ 0.029 & \textbf{0.518} $\pm$ 0.004 & 0.474 $\pm$ 0.013\\
    DAGMM   &  0.316 $\pm$ 0.000 & 0.316 $\pm$ 0.013 & 0.365 $\pm$ 0.020 & 0.362 $\pm$ 0.015 & 0.372 $\pm$ 0.012 \\
    DSEBMs  & \textbf{0.763}  $\pm$ 0.012 & \textbf{0.697} $\pm$ 0.011 & \textbf{0.666} $\pm$ 0.007 & 0.515 $\pm$  0.003 & 0.473 $\pm$ 0.003 \\
    IF      &  	0.368   & 	 0.372    & 0.365	& 0.301 &	0.298\\
    LOF     &  0.580	   & 0.438	     & 0.421	& 0.498 & \textbf{0.486}	\\
    OCGAN   &  0.408 $\pm$ 0.045 & 0.334 $\pm$ 0.098 & 0.365 $\pm$ 0.106 & 0.504 $\pm$ 0.083 & 0.497 $\pm$ 0.094\\
    OCSVM   &  	0.746   & 	0.681     & 0.637	& 0.467 &0.438\\
    RSRAE   &  0.593 $\pm$ 0.051 & 0.563 $\pm$ 0.035 & 0.488 $\pm$ 0.036 &  0.403$\pm$ 0.001 & 0.415 $\pm$ 0.003\\
    \bottomrule
  \end{tabular}}
\end{table}

%%%%%%%%%%%%%%%%%%%%%%%%%%%%%%%%%%%%%%%%%%%%%%%%%%%%

%\newpage
%\subsection{Table representation for Fig.~\ref{fig:variations}}\label{subsec:table_var}

\begin{table}[h!]
  \caption{AUC scores of KDD-99 for variations of MAW}
  \label{tab:var_kdd_auc}
  \centering
  \resizebox{0.67\columnwidth}{!}{
  \begin{tabular}{lccccc}
    \toprule
    \multicolumn{1}{}{} &\multicolumn{5}{c}{Training ratio of outliers per inliers, $c$} \\
    \cmidrule(r){2-6}
    Methods & 0.1 & 0.2 & 0.3 & 0.4 & 0.5 \\
    \midrule
    MAW    & \textbf{0.945} $\pm$ 0.028 & \textbf{0.906} $\pm$ 0.018 & \textbf{0.832} $\pm$ 0.016 & \textbf{0.775} $\pm$ 0.023 & \textbf{0.731} $\pm$ 0.017	\\
    MAW-MSE    & 0.844 $\pm$ 0.039 & 0.812 $\pm$ 0.032 & 0.746 $\pm$ 0.044 & 0.709 $\pm$ 0.020 & 0.675 $\pm$ 0.014\\
    MAW-KL divergence  & 0.905 $\pm$ 0.026& 0.863 $\pm$ 0.028&  0.801 $\pm$ 0.029 & 0.752 $\pm$ 0.016 & 0.696 $\pm$ 0.018	\\
    MAW-same rank     & 0.912 $\pm$ 0.023 & 0.868 $\pm$ 0.011& 0.797 $\pm$ 0.022 &0.750  $\pm$ 0.012 &0.699 $\pm$	0.040\\
    MAW-single Gaussian   &  0.914 $\pm$ 0.016 & 0.862 $\pm$ 0.021& 0.796 $\pm$ 0.013&0.751 $\pm$ 0.040 & 0.701 $\pm$ 0.045\\
    MAW-diagonal cov.  & 0.918 $\pm$ 0.023 & 0.858 $\pm$ 0.020& 0.801  $\pm$ 0.044& 0.743 $\pm$ 0.017& 0.703 $\pm$ 0.015\\
    VAE   & 0.821 $\pm$ 0.048 & 0.785 $\pm$ 0.027 & 0.732 $\pm$ 0.046& 0.717 $\pm$ 0.018 &0.685 $\pm$ 0.027 \\
    GMM prior   & 0.677 $\pm$ 0.009&  0.635$\pm$0.013 & 0.604 $\pm$ 0.006& 0.562 $\pm$ 0.015 & 0.517$\pm$0.014\\
    \bottomrule
  \end{tabular}}
\end{table}

\begin{table}[h!]
  \caption{AP scores of KDDCUP-99 for variations of MAW}
  \label{tab:var_kdd_ap}
  \centering
  \resizebox{0.67\columnwidth}{!}{
  \begin{tabular}{lccccc}
    \toprule
    \multicolumn{1}{}{} &\multicolumn{5}{c}{Training ratio of outliers per inliers, $c$} \\
    \cmidrule(r){2-6}
    Methods & 0.1 & 0.2 & 0.3 & 0.4 & 0.5 \\
    \midrule
    MAW    & \textbf{0.765} $\pm$ 0.025 & \textbf{0.732} $\pm$ 0.015 & \textbf{0.647} $\pm$ 0.012& \textbf{0.594}$\pm$ 0.014 &\textbf{0.556} $\pm$ 0.014\\
    MAW-MSE    & 0.715 $\pm$ 0.079 & 0.589 $\pm$ 0.058& 0.524 $\pm$ 0.053& 0.463$\pm$0.042 &0.410 $\pm$ 0.028\\
    MAW-KL divergence  & 0.735 $\pm$ 0.028 & 0.676 $\pm$ 0.028&  0.618 $\pm$ 0.024& 0.579$\pm$0.023 & 0.509$\pm$0.017\\
    MAW-same rank     & 0.725 $\pm$ 0.028& 0.681 $\pm$0.015 &0.622  $\pm$ 0.024&0.572 $\pm$0.017 &0.532 $\pm$0.038\\
    MAW-single Gaussian   & 0.737 $\pm$ 0.018& 0.675 $\pm$0.023 & 0.620 $\pm$ 0.025 & 0.569$\pm$0.036 &0.519 $\pm$0.044\\
    MAW-diagonal cov.  & 0.724 $\pm$ 0.021& 0.678 $\pm$ 0.035 &  0.589 $\pm$ 0.064&0.546 $\pm$0.019 &0.512 $\pm$0.016\\
    VAE   & 0.642 $\pm$ 0.030& 0.555 $\pm$ 0.043 & 0.524 $\pm$ 0.028& 0.478$\pm$0.024 & 0.450$\pm$0.015\\
    GMM prior   & 0.604 $\pm$ 0.010&  0.587$\pm$0.013 & 0.557 $\pm$ 0.009 & 0.534 $\pm$ 0.011 & 0.501$\pm$0.015\\
    \bottomrule
  \end{tabular}}
\end{table}

\begin{table}[h!]
  \caption{AUC scores of COVID-19 for variations of MAW}
  \label{tab:var_cov_auc}
  \centering
  \resizebox{0.67\columnwidth}{!}{
  \begin{tabular}{lccccc}
    \toprule
    \multicolumn{1}{}{} &\multicolumn{5}{c}{Training ratio of outliers per inliers, $c$} \\
    \cmidrule(r){2-6}
    Methods & 0.1 & 0.2 & 0.3 & 0.4 & 0.5 \\
    \midrule
    MAW    & \textbf{0.652} $\pm$ 0.021 & \textbf{0.609} $\pm$ 0.018 & \textbf{0.576} $\pm$ 0.019&\textbf{0.531} $\pm$0.020 & \textbf{0.504}$\pm$0.010	\\
    MAW-MSE     & 0.602 $\pm$ 0.022& 0.554 $\pm$0.063 &0.528  $\pm$ 0.041& 0.507$\pm$ 0.014& 0.479$\pm$0.021\\
    MAW-KL divergence  & 0.614 $\pm$ 0.025 & 0.580 $\pm$ 0.026 & 0.508  $\pm$ 0.064&0.476 $\pm$ 0.023& 0.463$\pm$0.016\\
    MAW-same rank     &  0.604 $\pm$ 0.031& 0.574 $\pm$ 0.048& 0.527 $\pm$ 0.044&0.430 $\pm$ 0.017& 0.408$\pm$0.021\\
    MAW-single Gaussian   & 0.621 $\pm$ 0.027&  0.586$\pm$0.029 &0.507  $\pm$ 0.047& 0.492$\pm$0.021 & 0.472$\pm$0.019\\
    MAW-diagonal cov.  & 0.600 $\pm$ 0.029&  0.586$\pm$ 0.030 & 0.535 $\pm$ 0.035&0.446 $\pm$0.028 &0.439 $\pm$0.038\\
    VAE   & 0.619 $\pm$ 0.073&  0.565$\pm$0.065 & 0.522 $\pm$ 0.049& 0.508 $\pm$ 0.023 & 0.473$\pm$0.016\\
    GMM prior   & 0.548 $\pm$ 0.012&  0.514$\pm$0.008 & 0.489 $\pm$ 0.010& 0.476 $\pm$ 0.011 & 0.469$\pm$0.009\\
    \bottomrule
  \end{tabular}}
\end{table}

\begin{table}[h!]
  \caption{AP scores of COVID-19 for variations of MAW}
  \label{tab:var_cov_ap}
  \centering
  \resizebox{0.67\columnwidth}{!}{
  \begin{tabular}{lccccc}
    \toprule
    \multicolumn{1}{}{} &\multicolumn{5}{c}{Training ratio of outliers per inliers, $c$} \\
    \cmidrule(r){2-6}
    Methods & 0.1 & 0.2 & 0.3 & 0.4 & 0.5 \\
    \midrule
    MAW    & \textbf{0.459} $\pm$ 0.014 & \textbf{0.442 } $\pm$ 0.011 & \textbf{0.424} $\pm$	0.018& \textbf{0.368}$\pm$0.015 & \textbf{0.353}$\pm$0.013\\
    MAW-MSE   & 0.421 $\pm$ 0.015 & 0.395 $\pm$ 0.025& 0.377 $\pm$ 0.012 & 0.332$\pm$0.013 &0.328 $\pm$	0.020\\
    MAW-KL divergence & 0.427 $\pm$ 0.016&  0.403$\pm$ 0.012 & 0.370  $\pm$ 0.021& 0.322$\pm$0.017 &0.313 $\pm$0.013	\\
    MAW-same rank     & 0.422 $\pm$ 0.021& 0.413 $\pm$0.026 & 0.375 $\pm$ 0.019  &0.344 $\pm$0.023 &0.335 $\pm$0.017	\\
    MAW-single Gaussian   &  0.425$\pm$ 0.019& 0.409 $\pm$ 0.012 & 0.374 $\pm$ 0.016& 0.339$\pm$ 0.014& 0.329$\pm$0.016\\
    MAW-diagonal cov.  & 0.412 $\pm$ 0.016& 0.397 $\pm$ 0.018&  0.369 $\pm$ 0.012& 0.343$\pm$0.009 &0.330 $\pm$0.009\\
    VAE   & 0.412 $\pm$ 0.030&  0.411 $\pm$ 0.043  & 0.379 $\pm$ 0.028& 0.341$\pm$0.011 & 0.333$\pm$0.013 \\
    GMM prior   & 0.389 $\pm$ 0.009 &  0.382$\pm$0.006 & 0.362 $\pm$ 0.005 & 0.346 $\pm$ 0.009 & 0.309 $\pm$0.0096\\
    \bottomrule
  \end{tabular}}
\end{table}

\end{document}